\documentclass[preprint,12pt]{elsarticle}
\usepackage[margin=2.5cm]{geometry}
\usepackage{ragged2e}
\usepackage{amsmath}
\usepackage[colorlinks=true,linkcolor=blue]{hyperref}
\usepackage[dvipsnames]{xcolor}
\usepackage{xurl}
\usepackage{chngcntr}
\usepackage{makecell}
\usepackage{courier}

\usepackage{lineno}
\usepackage{amssymb}
\usepackage{amsthm}
\usepackage{longtable}
\usepackage{threeparttable}
\usepackage{multirow} 
\usepackage{setspace}

\journal{journal}

\begin{document}
%\doublespacing

\begin{frontmatter}

\title{The causal relation between off-street parking and electric vehicle adoption in Scotland}

\author[inst1,inst3,inst4]{Bernardino D'Amico\corref{cor1}}
\ead{B.DAmico@napier.ac.uk}
\author[inst2,inst4]{Achille Fonzone}
\ead{A.Fonzone@napier.ac.uk}
\author[inst3,inst4]{Emma Hart}
\ead{E.Hart@napier.ac.uk}
\cortext[cor1]{Corresponding author}

\affiliation[inst1]{organization={Digital Built Environment Research Group (DiBEG)}, country={UK}}
\affiliation[inst2]{organization={Transport Research Institute}, country={UK}}
\affiliation[inst3]{organization={Centre For Artificial Intelligence and Robotic}, country={UK}}
\affiliation[inst4]{organization={School of Engineering and the Built Environment, Edinburgh Napier University, 10 Colinton Road, Edinburgh EH10 5DT}, country={UK}}

\begin{abstract}
The transition to electric mobility depends on widespread adoption and equitable access. We examine whether the “charging divide” between households with and without off-street parking reflects a true infrastructure constraint or underlying socio-economic disparities. We apply a probabilistic graphical model to nationally representative Scottish household data to identify the causal effects of parking access and income on electric vehicles (EV) adoption. 
Results reveal that access to private off-street parking increases EV ownership probability from 3.3\% to 5.6\%, an estimated causal effect of +2.3 percentage points (70\% relative increase). The (1st; 99th) percentile bootstrap interval is $+0.26;+4.77$ percentage points, while the corresponding basic bootstrap interval is $-0.17;+4.34$ percentage points. The effect primarily operates by increasing conversion among households already intending to adopt, rather than expanding the pool of households considering adoption.
In contrast, income is identified as the primary barrier to market entry, with higher-income households showing a 23.1 percentage point lower probability of no intention to purchase EVs compared to lower-income households. 
Models optimised for predicting observational outcomes overstate the effect of parking by conflating correlation with causation due to confounding bias, \textcolor{Black}{specifically by inflating its impact among households with no intention to adopt an EV.}
\end{abstract}

\begin{keyword}
Electric vehicle \sep Causal ML \sep Bayesian networks 
\end{keyword}

\end{frontmatter}

\section{Introduction}

Despite the awareness that decarbonisation of mobility requires rethinking priorities among modes \cite{TransportScotland2021}, the transition to electric vehicles (EVs) remains a cornerstone of many government climate strategies, including the UK’s and Scotland’s legally binding commitments to net zero \cite{Hutton2025EV, ScottishGov2025ClimatePlan}. 
\textcolor{Black}{Beyond climate mitigation, EV policies are also increasingly embedded within broader industrial and geopolitical strategies, including efforts to secure positions in global automotive value chains and to reshape regional production networks through trade policy and investment incentives \cite{martinez2024beyond, dua2026china}.}

Recent empirical analyses and systematic reviews suggest that aggregate EV adoption rates are an insufficient indicator of a just transition, which requires equitable access to charging infrastructure across diverse communities in order to avoid reproducing existing socio-economic disparities \cite{varghese2024equitable, kuby2025ev}.

As the EV market expands beyond early adopters, current adoption patterns reveal a stark driveway `charging divide': off-street parking enables homeowners to install domestic chargers, thus acting as a primary driver for EV adoption. According to a recent survey report by the Electric Vehicle Association (EVA) \cite{evaengland2025steer}, 90\% of current EV owners in England have access to off-street parking, with 81\% of these utilising a home charger. 
This infrastructure gap may represent an important barrier: residents of multi-unit dwellings (e.g., blocks of flats) may face significant physical, financial, and regulatory hurdles (such as the requirement for third-party legal consents) that frequently preclude home-charging installation \cite{zhang2023electric, kuby2025ev}.
Beyond practical convenience, this may reflect an underlying financial disparity: according to EVA England, while 87\% of surveyed EV drivers with driveways report lower running costs compared with those of internal combustion engines, only 50\% of those without off-street parking access find their EVs cheaper to operate \cite{evaengland2025steer}. 

\subsection{Determinants and barriers to EV adoption}
\label{sec:Determinants_EV_adoption}
According to existing literature, households' EV adoption is shaped by a combination of socio-demographic, economic, behavioural, and infrastructural factors. Empirical studies consistently identify income \cite{farkas2018environmental, haustein2018factors, chen2020assessing, trommer2015early, bruckmann2021battery, zhang2025electrifying, halse2025local}, education \cite{farkas2018environmental, haustein2018factors, trommer2015early, mukherjee2020factors}, home ownership \cite{hajhashemi2024modelling, bruckmann2021battery, jenn2020depth}, and multi-vehicle household status \cite{haustein2018factors, bruckmann2021battery, nazari2024electric} as strong predictors of adoption, reflecting both financial capacity and lifestyle compatibility with electric mobility. Psychological factors such as environmental concern \cite{trommer2015early, hajhashemi2024modelling} and technology affinity \cite{trommer2015early, bruckmann2021battery, salari2022electric, song2025exploring} also seem to influence adoption decisions. 
More recently, Budnitz et al. \cite{budnitz2026electric} applied the Theory of Planned Behaviour to UK residents parking in shared and public spaces, suggesting that internal psychological constructs, such as personal norms and attitudes, may exert a stronger influence on adoption intentions than the perceived ease of access to charging infrastructure itself.

Among structural determinants, housing characteristics and access to residential charging infrastructure have also been recognised as particularly influential. In line with Eva England statistics mentioned above, access to home charging capability is one of the strongest covariates \cite{moradloo2024charging}, explaining why detached housing and dedicated parking consistently predict ownership \cite{trommer2015early}. 
Experimental survey data analyses suggest that when the technical and financial `friction' of charging is hypothetically removed, `readiness' to adopt EVs nearly doubles among those with guaranteed private access \cite{patt2019availability}.
Home charging capability is also reported to be a more important determinant of EV adoption than the availability of public charging infrastructure \cite{bruckmann2021battery}.
Furthermore, spatial analyses have highlighted the joint effects of infrastructure, suggesting that while access to private parking is the primary driver, workplace charging can serve as a critical secondary support that partially mitigates the lack of home charging for some cohorts \cite{funke2019much}.
\textcolor{Black}{Importantly, these infrastructural constraints may influence not only first-time adoption but also long-term retention within the EV market, as recent evidence suggests that limited charging access remains a key factor driving EV discontinuance and repurchase decisions \cite{dua2024american, hardman2021understanding}.}

\subsection{Research gap}
\label{sec:research_gap}
Although the above-reviewed literature consistently reports the existence of a statistical association between dedicated parking provision and EV uptake, a significant gap remains in how we interpret and quantify this link. 
Two of the most relevant existing contributions approach the issue from distinct methodological angles. Patt et al. \cite{patt2019availability} rely on a randomised controlled survey design to bypass the inherent limitations of inferring causal effects from observational data, while Budnitz et al. \cite{budnitz2026electric} utilise Structural Equation Modelling (SEM) to examine the psychological drivers and latent intentions behind adoption. While both provide valuable insights into consumer likelihood to purchase an EV, this study identifies two critical voids in the current literature. 

First, there is a fundamental dimensionality gap in how adoption is measured. Existing studies, most notably Patt et al. \cite{patt2019availability}, employ one-dimensional (e.g. Likert) scales to measure the likelihood of purchase. While useful for gauging sentiment, such metrics collapse the complex transition from market non-participation to vehicle ownership into a single value, potentially obscuring where in the process a barrier actually operates. By contrast, we propose a multi-stage `adoption pipeline', categorising households across four states: no intention to buy, future intent, near-term intent, and actual ownership. This allows us to distinguish between factors that act as a market gatekeeper (influencing the initial decision to consider an EV) and those associated with higher likelihood of being in an advanced stage of the adoption pipeline. 
\textcolor{Black}{Importantly, these categories represent ordered states of adoption readiness along a behavioural continuum, ranging from non-consideration to active ownership, with the intermediate categories (``thinking to buy in the future'' and ``thinking to buy soon'') reflecting progressively shorter self-reported planning horizons toward adoption, while ``already own an EV'' represents realised adoption. By contrast, the ``not considering'' category does not encode a temporal horizon, but instead reflects a qualitatively distinct state of non-engagement within the decision process. The adoption pipeline is used as an analytical construct here to structure heterogeneity in EV adoption propensity, rather than a longitudinally observed process.}

Second, there remains a lack of causal evidence derived from revealed, real-world behaviour. Patt et al. \cite{patt2019availability} explicitly argue that using actual purchase data suffers from `serious shortcomings' because it is nearly impossible to statistically separate infrastructure access from confounding variables such as household wealth, multi-car ownership, and suburban driving patterns. Consequently, prior research investigating the specific effect of residential parking constraints has often avoided observational data, opting instead for stated-preference scenarios where variables' confounding is `solved' by stated-preference survey experiment \cite{patt2019availability}. However, such approaches risk introducing systematic `hypothetical' bias, whereby stated preferences diverge from actual behaviour due to the absence of real-world constraints \cite{haghani2021hypothetical}. 

Such reliance on stated-preference data for off-street parking stands in contrast to the wider empirical EV adoption literature, where researchers have successfully leveraged observational data to estimate causal effects for other determinants, most notably, the provision of public charging infrastructure. In that domain, a robust body of applied econometrics utilises Instrumental Variable (IV) designs to isolate the causal impact of public charging infrastructure deployment on local EV uptake, e.g., Springel (2021) \cite{springel2021network}, Li et al. (2017) \cite{li2017market}.

Our approach departs from stated-preference by confronting the socio-economic entanglement that has historically rendered revealed behavioural data `unusable' for causal analysis. 
In doing so, we directly confront the problem of treatment endogeneity (a challenge traditionally addressed in econometrics through quasi-experimental approaches such as IV \cite{angrist1996identification}). Within the broader framework of causal graphical models, IV can be understood as a specific identification strategy in which validity is characterised by graphical conditions that ensure exogenous variation in the treatment in the presence of unobserved confounding. We specify a Causal Bayesian Network to encode the system-wide structural assumptions governing the system and identify admissible adjustment sets via graphical criteria. Under these assumptions, we estimate the causal effect of parking provision by adjusting for observed confounders while explicitly formalising assumptions regarding latent confounding that may bias observational estimates.

To this end, although the absence of off-street parking may constitute a structural constraint on home charging, it remains unclear to what extent the observed adoption gap reflects the infrastructural barrier itself rather than the socio-economic characteristics that systematically correlate with a lack of off-street parking. Disentangling these effects is essential for effective policy design. This leaves a critical policy question unresolved: is the lower adoption rate among on-street parking households primarily driven by charging access constraints, or by broader socio-economic disparities that shape both vehicle ownership and off-street parking availability?
International evidence indicates that affluent and lower-density areas are disproportionately favoured in charging station deployment, generating a `spatial disparity' that overlaps with existing demographic privilege \cite{ermagun2024charging}. Historically, UK strategies have prioritised vehicle purchase incentives; however, if the principal barrier faced by the 40\% of UK households without driveway access \cite{Binns2025} lies in charging convenience rather than upfront vehicle cost, a reorientation of policy towards sustainable and equitable infrastructure provision may be required.

Crucially, policies supporting residential charging must navigate a trade-off: incentivising parking-based infrastructure may conflict with urban planning goals aimed at increasing density and promoting mixed land use to reduce sprawl. We therefore clarify that this study does not evaluate the expansion of off-street parking as a viable policy lever. Such an intervention would be physically infeasible in dense urban areas and counterproductive to broader decarbonisation objectives when applied to new housing developments. Following the logic of causal inference for non-manipulable causes, we treat off-street parking provision as a structural diagnostic rather than a literal policy lever. Just as quantifying the causal link between obesity and morbidity helps target taxation policies at sugary drinks rather than mandating weight loss itself \cite{pearl2018does}, estimating the ‘driveway effect’ can help calibrate compensatory policy interventions aimed at replicating the advantages that driveway access confers. In practice, these advantages operate through two primary mechanisms: a price mechanism, whereby domestic charging enables access to cheaper electricity tariffs compared with public charging; and a temporal mechanism, which enables a transition from high-friction active charging to passive, zero-wait-time background activity. While we do not model these mechanisms explicitly, identifying the causal effect of parking provision provides a useful proxy for the scale of cost reductions and convenience improvements required to bridge any existing adoption gap without compromising urban density.

 \begin{figure}[ht!]
\centering
  \includegraphics[width=6.5cm]{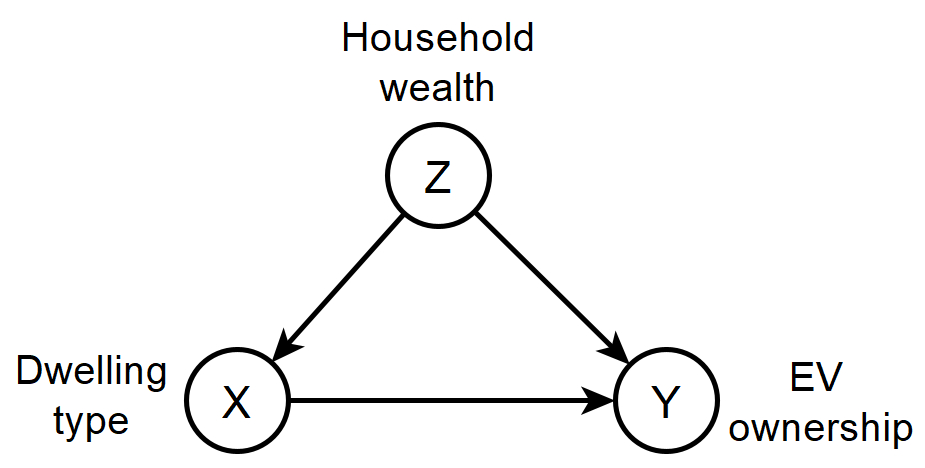}
  \caption{Causal graphical structure of the parking provision and EV ownership toy example. The presence of household wealth ($Z$) creates an undirected back-door path ($X \leftarrow Z \rightarrow Y$), inducing a spurious correlation between dwelling type ($X$) and EV ownership ($Y$) that biases the true causal effect of interest, $X \rightarrow Y$.}
  \label{fig:backdoor_toy}
\end{figure}

\subsection{The necessity of causal inference for policy analysis}
\label{section:WhyCausal}

In this context, Scotland has been piloting one-off schemes to address the off-street charging gap, such as the Energy Saving Trust’s cross‑pavement charging initiative, which facilitates the installation of cable gullies for households reliant on on-street parking \cite{EnergySavingTrust2024, browne2025crosspavement}. In parallel, regulatory levers are also being considered as additional policy avenues, including the possibility of removing existing restrictions to expand on‑street and pavement charging as part of permitted development rights \cite{scotgov2023pdrphase2}. 

In order to ensure such targeted policy interventions are effective, policymakers must move beyond standard correlations and apply a causal framework capable of disentangling physical infrastructure from socio-economic privilege. To this end, the primary aim of this research is to quantify the causal effect of private off-street parking on EV adoption in Scotland, moving beyond traditional correlational analyses to address a critical void in the literature.

In research domains concerning policy analysis and evaluation (or decision making in general), the objective shifts from predicting observations to predicting the effects of interventions. While machine learning excels at estimating conditional probabilities, such as the likelihood of EV ownership given a set of covariates, these models are fundamentally optimised for statistical association rather than causal identification \cite{d2025cuts, DAmico2026}. As argued by Chauhan et al. in the context of travel mode choice \cite{chauhan2024determining}, even high-accuracy predictive models (e.g., based on deep neural networks or random forests) may succeed in capturing complex data patterns while failing to reflect the underlying causal mechanisms required for robust policy design. This discrepancy arises from the critical distinction between observing a state and intervening upon it.

To illustrate, let us consider the relationship between dwelling type ($X$) and EV ownership ($Y$). A purely predictive model might find a strong positive association between the two, as a detached house is significantly more likely to have a private driveway than an apartment. However, household wealth ($Z$) may act as a critical confounder, causally influencing both the capacity to afford a dwelling with a driveway ($X$) and the financial ability to purchase an EV ($Y$). In causal graphical terms (Figure \ref{fig:backdoor_toy}), the variable $Z$ creates an undirected `back-door' path ($X \leftarrow Z \rightarrow Y$), which induces a spurious correlation between $X$ and $Y$ \cite{pearl2009causality}. Without `closing' this path by conditioning on $Z$, any observed difference in EV ownership between households with and without off-street parking will conflate the true causal effect with a bias driven by underlying economic status.

\textcolor{Black}{This distinction between observational and interventional reasoning is formalised using Pearl’s do-operator $do(\cdot)$, whereby $P(Y \mid do(X=x))$ represents the probability distribution of $Y$ under an external intervention that fixes $X$ to $x$, thereby removing the influence of its upstream causes in the causal graph. In general, $P(Y \mid do(X=x)) \neq P(Y \mid X=x)$ in the presence of confounding.}

\subsection{Research questions and causal estimands}
\label{section:ResearchQuestions}

Building on the distinction between observational associations and interventional effects, we formalise our policy problem in explicitly causal terms. Rather than asking whether off-street parking or income is \emph{associated} with EV adoption, we evaluate the shift in outcomes under hypothetical interventions. Given that the relationship between parking and adoption is fundamentally entangled with socio-economic status, we formulate three primary research questions to disentangle these drivers:

\begin{itemize}
\item[RQ1:] What is the causal effect of providing access to off-street parking on EV adoption status and purchase intentions?
\item[RQ2:] How does household income influence the EV adoption pipeline, acting both as a direct driver of affordability and as a primary determinant of residential parking access?
\item[RQ3:] To what extent does the observational difference in EV adoption probabilities diverge from the actual probability shift under intervention, and what specific policy bias results from mistaking the former for the latter?
\end{itemize}
We translate the first two questions into formal causal estimands, whilst the third evaluates the discrepancy between interventional and standard observational probabilities. Together, this framework enables us to determine whether the provision (or absence) of off-street parking functions as a driver (or barrier) within the EV adoption pipeline; whether household income represents a more fundamental prerequisite for market entry; and to what extent purely observational metrics may misdirect policy interventions aimed at facilitating the transition to electric mobility.

\section{Methods}
\label{sec_methods}
Our methodology is grounded in probabilistic causation \cite{suppes1973probabilistic}, whereby external interventions are understood as factors that shift the likelihood of an outcome rather than as deterministic mechanisms. Specifically, we employ a causal Bayesian network framework \cite{pearl2009causality}, representing the system as a directed acyclic graph (DAG), such as the one illustrated in Figure \ref{fig:backdoor_toy}.

The topology of such graphs, denoted by $\mathcal{G}$, can be established through two primary approaches. The first is knowledge-based specification, in which causal relationships are defined based on prior domain expertise \cite{tennant2021use}. The second approach is automated structure learning, whereby the graph is inferred directly from observational data. In this study, we adopt a multi-stage hybrid strategy. First, domain expertise is exercised in the selection of the feature space; by determining which variables are included in the model, we encode prior knowledge regarding the known socio-economic and infrastructural determinants of EV adoption. Second, we learn the causal structure from data while enforcing domain-specific constraints during the automated search process, followed by a post-hoc manual refinement of the resulting graphical output. The algorithm details and refinement steps are described in section \ref{section:CausalStructureLearning}.

In a causal Bayesian network, nodes represent random variables, and directed edges encode direct causal relationships. As for non-causal Bayesian networks, each observed variable $V_i \in \boldsymbol{V}$ is associated with a conditional probability distribution, which is denoted as $P(V_i = v_i \mid \boldsymbol{Pa}(V_i))_\mathcal{G}$, and specifies the probability that $V_i = v_i$ given the values of its parent\footnote{Parents are the set of nodes in $\mathcal{G}$ from which a directed edge points toward $V_i$.} variables $\boldsymbol{Pa}(V_i)$ in the graph $\mathcal{G}$. The implementation of Bayesian networks relies on two fundamental assumptions that connect the graphical structure to the observed statistical data:

\begin{itemize}
\item[-] Faithfulness condition: it assumes that every statistical independence found in the data is a direct result of the underlying data-generating process, as modelled by the graph’s structure, therefore assuming that different causal influences do not perfectly cancel each other out across different pathways.

\item[-] Causal Markov condition: this assumption states that, conditional on its direct parents in the causal graph, each variable is statistically independent of all its non-descendants. In essence, once the immediate causes (i.e., parents) of a variable are accounted for, no additional information from its ancestors or other non-descendant nodes contributes further to its probability belief.
\end{itemize}

To illustrate the latter with an example, consider the causal chain: weather ($X$) $\rightarrow$ traffic ($Z$) $\rightarrow$ being late ($Y$). In this causal chain, the Markov condition implies that if we already know the current state of traffic ($Z=true$), knowing the weather provides no additional information about the probability of being late. The causal influence of weather $X=rainy$ on arrival time is entirely captured by the traffic congestion it creates, hence, once the state of traffic is observed, knowing the weather becomes redundant for predicting lateness:  $P(Y \mid Z, X) = P(Y \mid Z)$, thus implying that $Y \perp X \mid Z$, meaning, the outcome $Y$ is independent of its `grandparent(s)' $X$ once we condition on its parent variable(s) $Z$. 

A critical implication of the causal Markov condition is the possibility to factorise the full joint probability distribution over the entire set $\mathbf{V}$ of observed variables. In the absence of a structured graph, representing the full joint distribution explicitly would require specifying dependencies across the entire variables' state space, yet as the number of variables increases, this task becomes computationally and statistically intractable due to the curse of dimensionality, that is, the exponential blow-up in the degrees of freedom required to represent the full joint.\footnote{For the specific variable set $\mathbf{V}$ used in this study ($n=15$), an explicit representation of the full joint distribution would require specifying 15,482,880,000 distinct joint state configurations, computed as the product of the cardinalities of all variables’ state spaces: $\prod_{i=1}^{15} |V_i|$}
Conversely, in a Bayesian network, the probability of a generic event assignment $\{V_1=v_1, V_2=v_2, \dots, V_n=v_n\}$ can be implicitly encoded as a product of local conditional probabilities \cite{koller2009probabilistic}:

\begin{equation}
    \label{Bayesian_factorisation}
    P(v_1, v_2,...,v_n)_{\mathcal{G}}= \prod_{i=1}^{n} P(v_i \mid \boldsymbol{Pa}(V_i))_\mathcal{G}
\end{equation}
As such, any observational probability distribution (whether marginal or conditional on observing a set of covariates) can be queried from the network's local probability distributions via Eq. (\ref{Bayesian_factorisation}).

\subsection{Parameter estimation}
Once the DAG structure $\mathcal{G}$ is specified (see section \ref{section:CausalStructureLearning}), the model can then be trained to represent the specific joint distribution of the system (a procedure known as parameter estimation), which involves quantifying the local conditional probability distributions on the r.h.s. of Eq. (\ref{Bayesian_factorisation}). For discrete variables, as used in this study, these local distributions can be represented via Conditional Probability Tables (CPTs), hence training the network essentially entails a frequency estimation of each entry assignment in these tables. For each node $V_i$, the likelihood of each state $v_i$ conditioned on every possible configuration $\mathbf{pa}_j$ of its parent set was computed via Maximum Likelihood Estimation:
\begin{equation}
\label{eq:MLE}
P(v_i \mid \mathbf{pa}_j) = \frac{\text{count}(v_i, \mathbf{pa}_j) + \alpha}{\sum\limits_{v_i \in \mathcal{V}_i} [\text{count}(v_i, \mathbf{pa}_j) + \alpha]}
\end{equation}
where the numerator, $\text{count}(v_i, \mathbf{pa}_j)$, denotes the frequency of the joint occurrence of variable $V_i$ in state $v_i$ and of its parents in state $\mathbf{pa}_j$, whereas the denominator sums these counts across the entire set $\mathcal{V}_i$ of possible outcomes for the child variable $V_i$, thus acting as a normalisation factor to ensure that all probabilities within each row in the CPTs sum up to 1.
The Laplace smoothing term in Eq. (\ref{eq:MLE}) was set to $\alpha = 10^{-5}$. The term is required to address the ``zero-frequency'' problem, that is, when a parent-child state $\{v_i, \mathbf{pa}_j\}$ is not present in the training dataset, its count would be zero on the r.h.s. of Eq. (\ref{eq:MLE}), which would yield a zero probability for $P(v_i \mid \mathbf{pa}_j)$. 

Regarding notation, it is important to note that the subscript $v_i$ in the above summation in Eq. (\ref{eq:MLE}) is not a standard numerical index, but a placeholder for the elements of the set $\mathcal{V}_i$. For the sake of brevity we will henceforth adopt the shorthand $\sum_{v}$ to represent summations $\sum_{v \in \mathcal{V}}$ across all subsequent equations.

\subsection{Variable selection, data sourcing and preprocessing}
\label{sec:variableSelectionetc}
The selection of variables for our causal model followed a systematic, domain-informed approach. We began by defining the primary causal relationship of interest: the treatment and outcome variables. From this core link, we expanded the variable set by identifying factors documented in the literature as influencing both treatment and outcome. This expansion was guided by a synthesis of transport economics and urban sociology research. To support systematic literature mining, we used Elicit \cite{elicit}, an LLM-based research tool (AI assistant), to identify empirically grounded determinants in prior EV adoption studies (see literature review section \ref{sec:Determinants_EV_adoption} for a detailed summary). To ensure the technical validity of the generated outputs, we followed a verification protocol where each identified determinant was manually cross-referenced against the original source papers to confirm empirical significance and the direction of the reported effect. Only variables with a consensus of support across multiple studies were retained for the final model specification.

We idealised the system as comprising decision-making agents embedded within material and infrastructural constraints. Accordingly, the resulting variable set can be conceptually partitioned into two ontologically distinct categories: (1) socio-economic household characteristics, representing agent-level attributes such as net annual income, household composition, and working status; and (2) dwelling and infrastructure characteristics, representing structural and spatial constraints, including architectural dwelling type, dwelling age, and regional charging infrastructure density. 

This distinction between household characteristics and structural constraints is reflected in our data sources. 
To populate these variables with empirical observations, we integrated two complementary national datasets that together capture the intersection of social behaviour and physical housing conditions in Scotland, namely: the Scottish Household Survey (SHS) \cite{SHS2022} and the Scottish Housing Condition Survey (SHCS) \cite{SHCondS2022}. SHS and SHCS data were enriched with information about availability of public charging facilities as explained below.

The latest SHS and SHCS data releases at the time of writing were used for this study, referring to the 2022 survey year. The SHS contains approximately 10,500 household entries, reporting on a range of socio-economic characteristics and behaviours, including information indicating whether the selected random adult within the household has access to (or intends to buy) an electric car or van. 
\textcolor{Black}{While these data represent the most recent available records, EV prices, charging infrastructure, and public incentives have evolved since 2022, which may have some influence on the current interpretation of the estimated adoption pipeline presented in this analysis.}

Data from the SHS were cross-linked with those from the SHCS for the same year via a unique-household ID variable, mapping surveyed households to their corresponding dwelling characteristics. 
The latter represents a much smaller subset ($\approx$3,000 entries). While the SHS focuses on social and economic data, the SHCS provides an expert-led physical assessment of the Scottish housing stock, reporting on dwelling fabric, energy performance, and external property features.
By doing so, we were able to assemble a merged dataset reporting on both household and dwelling variables, such as dwelling type, dwelling age, and most importantly, parking provision. 
This latter variable, as originally reported in the SHCS, comprises nine separate categories. These were remapped into two categories (see Table \ref{tab:mapping_parking_provision} in \ref{sec_appendix_A}) to obtain a binary classification for our parking provision treatment variable: specifically, distinguishing between those who can park off-street and those who cannot.

Similar data cleaning and mapping procedures were applied to other variables. For instance, the SHC variable regarding EV ownership and intent included a category for individuals who stated they do not drive/need a vehicle, regardless of engine type; these respondents were subsequently excluded from the sample.

To prevent information loss during model discovery and training phases, the original 8-fold classification for household net annual income (termed $V_1$ in here ---see Table \ref{table:variable_list} in \ref{sec_appendix_A}) was retained in its original form. However, for causal estimation purpose the inferred post-intervention distributions $P( Y \mid do(V_1))$ \textcolor{Black}{---with $Y$ denoting the outcome variable representing EV adoption status and intentions (Table \ref{table:variable_list}, \ref{sec_appendix_A})}--- were later aggregated into two categories, namely: $V_1=\text{low-income}$ for bands 1 to 6 and $V_1=\text{high-income}$ for bands 7–8. 
This binary discretisation was achieved by averaging the specific post-intervention probability of each band, weighted by their respective marginal probabilities $P(v_1)$, (reported here in Table \ref{tab:mapping_income}) that is, for $V_1=\text{low-income}$ for instance we have:

\begin{equation}
P(Y \mid do(V_{1}=\text{low-inc.})) = \frac{\sum_{i=1}^{6} P(Y \mid do(v_{1,i})) P(v_{1,i})}{\sum_{i=1}^{6} P(v_{1,i})}
\end{equation}
By re-discretising $V_1$ approximately at the median (48.5\%), we obtain an almost perfectly balanced binary partition to contrast the effect of income between control group ($\leq$ £30,000) and treatment group (> £30,000).

The resulting full observational set, $\mathbf{V}= \{Y, V_1, \dots, V_{14} \}$, comprises 15 variables, the majority of which were retrieved from the SHS and the SHCS dataset collections, with the exceptions of $V_{10}$ and $V_{11}$, which represent the density of workplace and public EV charging stations, respectively. These two infrastructure density metrics were derived for each of Scotland's 32 local authorities by calculating the ratio between charging station projects completed by the end of 2022 and the total resident population within each jurisdiction. Charging station statistics were retrieved from the Department for Transport \cite{dft_charging_grants_2023, dft_public_charging_2025}, whereas the demographic data used to normalise these figures were sourced from the National Records of Scotland \cite{nrs_population_2024}. This real-valued density data was subsequently normalised per 100,000 residents and discretised into five ordinal bands, as detailed in Table \ref{table:variable_list} in \ref{sec_appendix_A}, before being cross-linked to individual household units in the primary training dataset using the \emph{Local Authority} code as a unique spatial identifier.

\subsubsection{Population weighting and resampling}
\label{sec:Population_weighting_resampling}
To ensure that the final model is representative of Scotland’s broader dwelling and household population, the unweighted dataset resulting from the aforementioned preprocessing and cleaning procedures was subjected to a weighted resampling routine. Specifically, the original sample ($N=1,802$) was resampled with replacement using selection probabilities proportional to the \emph{paired sampling weight} variable (\texttt{tsWghtP\_n}) provided by the SHCS. This procedure adjusts for the complex survey design by approximating a representative pseudo-population; by doing so, the structural dependencies and conditional probabilities learned by the network better reflect the population distribution across the weighted dimensions. While this approach mitigates the inherent biases of the sampling frame, we acknowledge that it corrects for non-response only relative to the auxiliary variables used in the weight construction. 
\textcolor{Black}{Furthermore, we acknowledge the presence of sampling uncertainty introduced by the limited sample size within certain strata (e.g., the low representation of EV-owning households in those strata). To account for this data sparsity problem, we employ a survey-weighted non-parametric bootstrap procedure (see section \ref{subsec:bootstrap_uncertainty}) to quantify the sensitivity of the estimated causal effects to sampling variability.}

\begin{table}[ht]
    \centering
    \begin{threeparttable}
    \caption{Marginal probabilities and cumulative sample distribution for the eight classification bands of household income ($V_1$). \label{tab:mapping_income}}
\begin{tabular}{|l|c|c|}
\hline
\textbf{\makecell[l]{Household  \\ income band, $v_1$}} & \textbf{\makecell[l]{Marginal  \\ probability $P(v_1)$}} & \textbf{\makecell[l]{Cumulative  \\ sample total}} \\ \hline
1: < £6000 & 0.0202 & 0.020 \\ 
2: £6001-10,000 & 0.0364 & 0.057 \\ 
3: £10,001-15,000 & 0.0786 & 0.135 \\ 
4: £15,001-20,000 & 0.1358 & 0.271 \\ 
5: £20,001-25,000 & 0.1079 & 0.379 \\ 
6: £25,001-30,000 & 0.1064 & 0.485 \\ 
7: £30,001-40,000 & 0.1602 & 0.646 \\ 
8: > £40,000 & 0.3546 & 1.000 \\ 
\hline
\end{tabular}
\end{threeparttable}
\end{table}

\subsection{Causal discovery}
\label{section:CausalStructureLearning}

Causal discovery aims to learn causal relationships among random variables directly from observational data. 
To discover the causal structure of our graphical model, we employed the Multivariate Information-based Inductive Causation (MIIC) framework \cite{verny2017learning, ribeiro2024learning}. MIIC was selected for its ability to handle the relatively limited sample size of our dataset. By quantifying the strength of dependencies rather than just their existence, MIIC effectively balances explanatory power with model complexity, outperforming classical constraint-based methods \cite{ribeiro2024learning}, particularly when dealing with small-sample datasets, where a high noise-to-signal ratio can undermine the reliability of statistical independence testing as typically employed in constraint-based methods. 
A technical note describing the theoretical foundations underpinning the MIIC algorithm, is detailed in  \ref{sec_appendix_B}.

Within this framework, the MIIC algorithm starts from a fully connected graph and iteratively prunes edges while orienting the remaining ones using information-theoretic metrics together with adaptive complexity thresholds. By expressing the complexity threshold as a ratio relative to the sample size, the method provides a formal mathematical implementation of Occam's Razor, penalising overly flexible models that may otherwise over-fit limited data. For instance, a fully connected graph can represent virtually any arbitrary distribution, thereby mistaking sampling noise for genuine structural dependencies.
Unlike traditional constraint-based approaches, which rely on fixed and often somewhat arbitrary significance thresholds (e.g., $p$-value < 0.005), the adaptive complexity threshold allows the algorithm to resolve increasingly subtle dependencies as the sample size grows, while maintaining structural parsimony in data-sparse regimes.

Following the pruning of the graph skeleton, the algorithm then orients the remaining edges and identifies potential latent confounding to produce a Maximal Ancestral Graph (MAG) by evaluating \emph{interaction} signatures and applying logical propagation rules, as detailed in \ref{subsection:edge_orient}.

\subsubsection{Blacklisting edges}
\label{sec:blacklisting}
To ensure logical consistency of the discovered structure, we imposed a set of a priori structural constraints during the iterative pruning and orientation phases. Specifically, we defined a `blacklist' of prohibited causal directions. For example, edges such as $Household\ income \rightarrow Household\ working \ status$ were blacklisted, as it is the occupational status that determines the flow of earnings rather than the reverse. 

\begin{figure}[b!]
\centering
  \includegraphics[width=16.5cm]{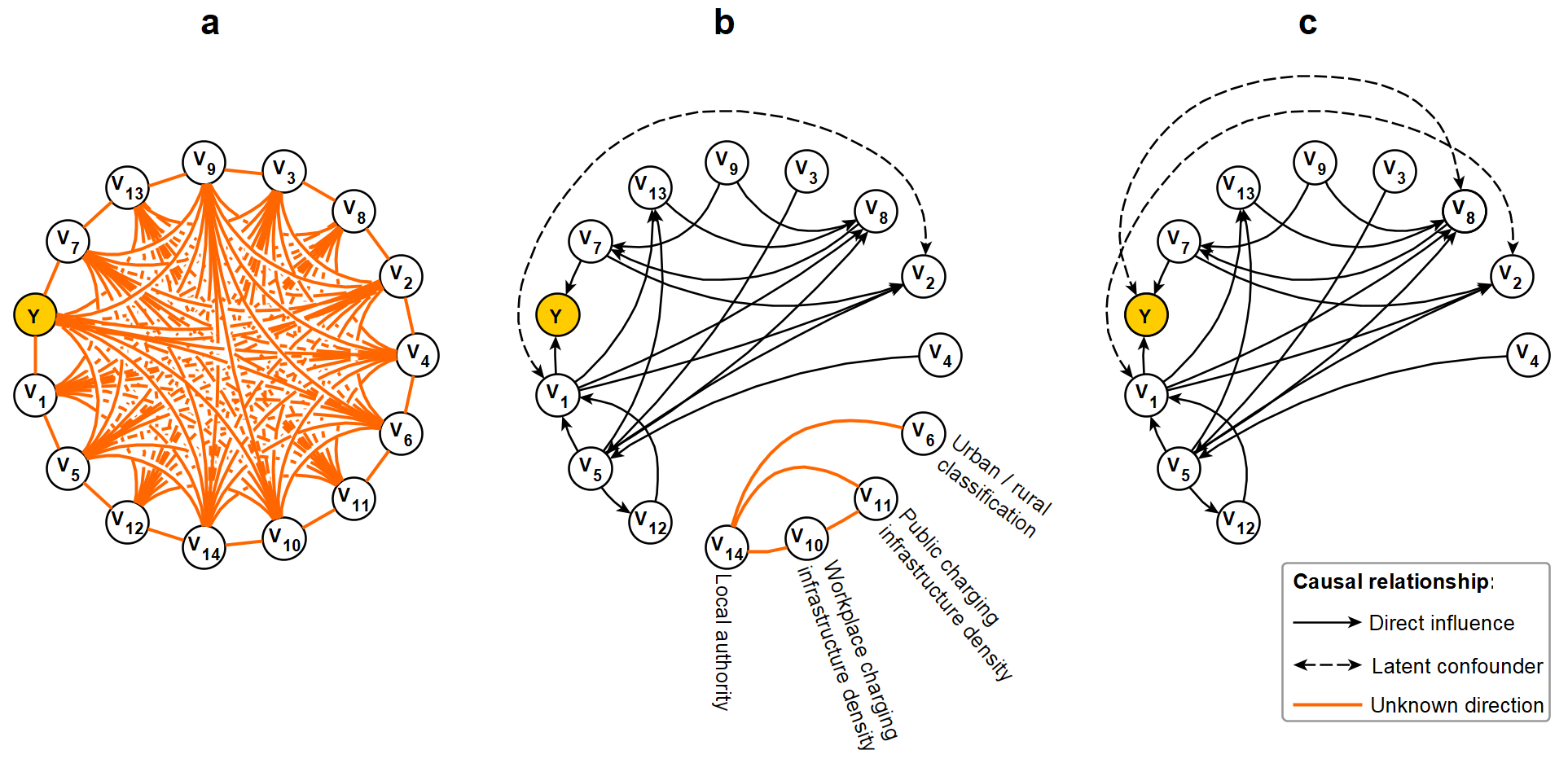}
  \caption{Causal discovery workflow. (a) Initial, fully connected skeleton. (b) Maximal Ancestral Graph discovered via the MIIC algorithm, featuring directed ($\rightarrow$), bi-directed ($\leftrightarrow$) and undirected ($-$) edges. The isolated variable subset $\{V_6, V_{10}, V_{11}, V_{14}\}$  is statistically independent of the main network.
(c) Final refined DAG: fully specified model after omitting the disconnected subset and incorporating post-hoc a latent confounder between $V_8$ and $Y$.}
  \label{fig:Causal_Discovery_workflow}
\end{figure}

\textcolor{Black}{Notably, the causal discovery algorithm did not detect causal links between public (workplace) infrastructure charging densities ($V_{11}$ and $V_{10}$, respectively) and individual EV ownership status \& intentions, $Y$ (see Figure \ref{fig:Causal_Discovery_workflow}-b). This was likely due to the coarse granularity of the infrastructure data, which was measured as the average density at the local authority level, i.e., count of public (workplace) charging stations per 100,000 residents. As such, the absence of detected edges should not be interpreted as evidence of no causal influence, but rather as a lack of identifiable signal at the resolution of the available data.}

\subsubsection{Post-hoc refinement}
\label{app_sec:post_hoc_refinement}
The automated discovery phase described above resulted in the MAG illustrated in Figure \ref{fig:Causal_Discovery_workflow}-b which provides a graphical representation of the Markov Equivalence Class, i.e., the class of all causal structures that are statistically indistinguishable based on the conditional independence relationships found in the data. Notably, all of the undirected edges the algorithm was unable to orient in the orientation-propagation phase (see \ref{subsection:edge_orient}) were confined to an isolated four-variable cluster (as shown in Figure \ref{fig:Causal_Discovery_workflow}-b), which shared no connectivity with the rest of the model, i.e., the two subgraphs are statistically independent. To this end, we were thus able to simply omit this four-variable subset from the final model, hence successfully recovering a fully specified DAG from the Markov Equivalence Class. 

\begin{figure}[ht!]
\centering
  \includegraphics[width=13cm]{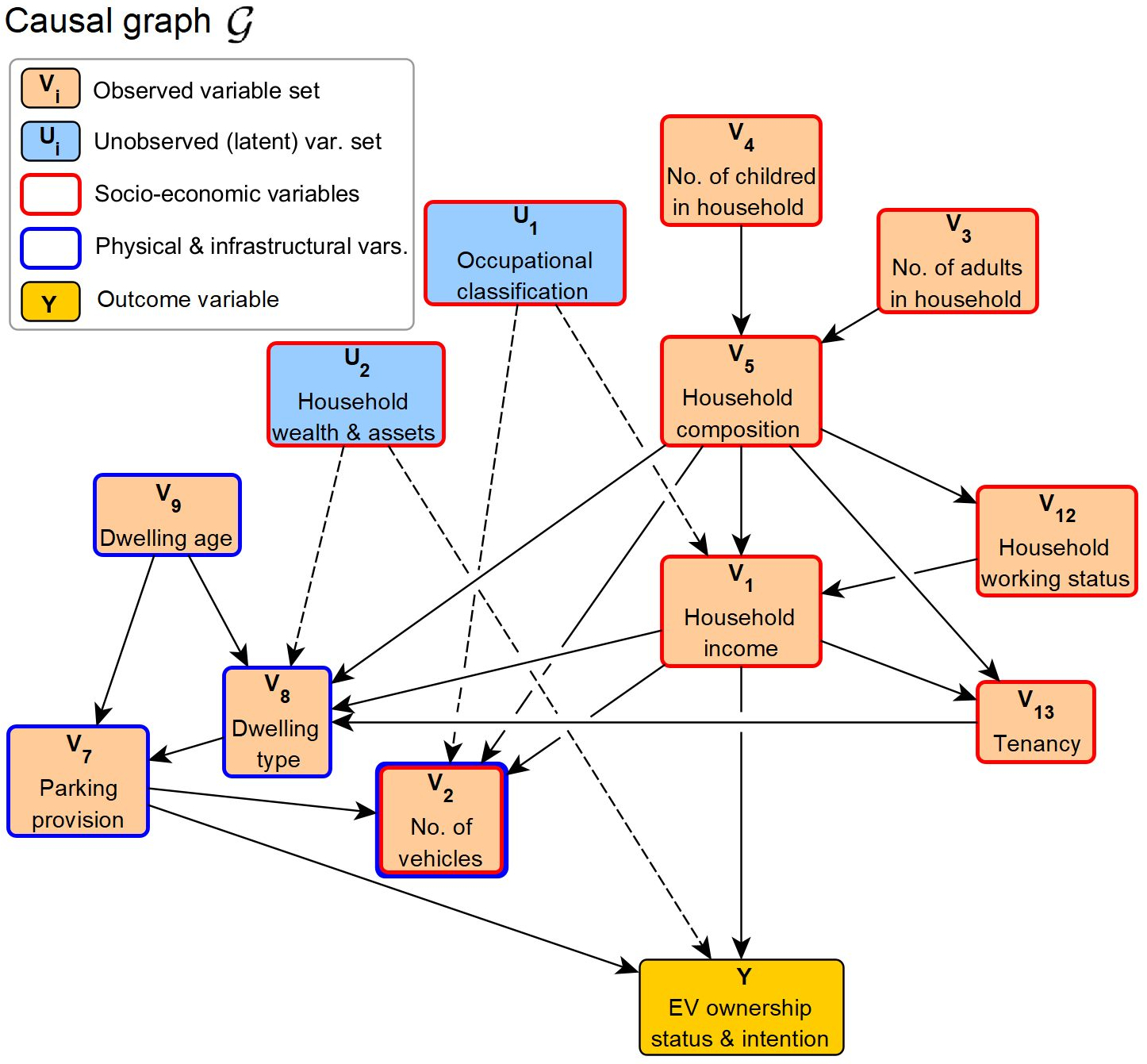}
  \caption{Final DAG ($\mathcal{G}$) representing the causal mechanisms of EV ownership status and intentions ($Y$). The model integrates automated causal discovery results with post-hoc manual refinements to account for latent confounding ($U_i$). Directed solid edges indicate causal effects, dashed edges denote unobserved common causes. The DAG structure provides the formal basis for identifying sufficient adjustment sets $\mathbf{Z}$ required to estimate post-treatment distributions from observational data via the back-door criterion (section \ref{section:Identification}).}
  \label{fig:final_dag}
\end{figure}

Additionally, a latent (unobserved) confounder was discovered by the algorithm between the \emph{household income} variable ($V_1$) and the variable reporting on \emph{No. of vehicles} ($V_2$), represented as a bi-directed edge in Figure \ref{fig:Causal_Discovery_workflow}-b. Because this dependency could not be explained by any other variables in our dataset, it implies the influence of an external factor not captured in our study. We hypothesise that this latent confounder likely captures the effect of occupational classification (denoted as $U_1$ in Figure \ref{fig:final_dag}), that is, specific work occupations inherently dictate both a household's income and its mobility requirements ---for example, trade-based occupations (e.g., self-employed contractors, electricians, or plumbers)--- thereby acting as a common parent to both variables.

A further bi-directed edge was manually introduced post-hoc between \emph{Dwelling type} ($V_8$) and \emph{EV ownership status \& intention} ($Y$) to account for the potential latent confounding effect of household's overall financial capacity. This latent factor, denoted as $U_2$ in Figure \ref{fig:final_dag}, represents accumulated wealth in the form of liquid assets and non-income financial reserves. We assume that both the choice of housing type and the purchase of premium vehicles, such as EVs, depend on a household’s overall financial capacity, including their savings and long-term wealth ($U_2$) as well as their annual income flow ($V_1$).
Although no bi-directed edge $V_8\leftrightarrow Y$ was detected via the MIIC algorithm, the absence of evidence does not imply evidence of absence: it is entirely possible that such confounding influences exist, yet the association $V_8 - Y$ might have been too weak (or the sample size too small) to surpass the complexity threshold leading the algorithm to treat it as statistically indistinguishable from noise.
\textcolor{Black}{By anchoring data-driven algorithmic discovery with domain-specific structural constraints and expert refinements, the final network topology $\mathcal{G}$ is subsequently treated as a fixed causal framework defined by domain theory.}

The final, manually refined, graph $\mathcal{G}$ (as shown in Figure \ref{fig:Causal_Discovery_workflow}-c and Figure \ref{fig:final_dag}) provides the necessary set up to perform causal effect identifiability and estimation.

\subsection{Causal effect identification \& estimation}
\label{section:Identification}

Having established the causal structure $\mathcal{G}$, we now focus on estimating the causal effect of specific interventions. Within the framework of causal graphical models, this amounts to computing the post-intervention distribution $P(Y \mid do(X))$. This quantity represents the distribution of the outcome $Y$ when the treatment variable $X$ is externally set to a specific value $x$. Graphically, such an intervention corresponds to removing all incoming edges to $X$, which encode the natural causal mechanisms that would otherwise determine its value.

As mentioned in the introduction, a central difficulty is that the post-intervention distribution $P(Y \mid do(X))$ generally differs from the observational conditional distribution $P(Y \mid X)$ available in the data ---with the latter reflecting how $Y$ behaves when we merely observe $X = x$, with all underlying causal mechanisms intact. In contrast, $P(Y \mid do(X))$ characterises the response of $Y$ when those mechanisms influencing $X$ have been deliberately severed through intervention.

This distinction naturally raises the question of \emph{identifiability}, that is, under what conditions can the interventional distribution be recovered from purely observational data.
An effect is said to be identifiable if the interventional distribution can be expressed entirely in terms of observational probabilities. In a Markovian model, i.e. where all variables are fully observed and no unmeasured confounding is present, such identification is in principle always possible from the joint observational distribution. 
However, this guarantee no longer holds when the graph contains latent variables, such as $U_1$ and $U_2$ in our case (so-called semi-Markovian case). Unobserved confounders may induce spurious associations between observed variables by creating unblocked back-door paths, and since these latent variables cannot be conditioned on directly (as we have no data for them), it may be impossible to express the interventional distribution solely in terms of observational quantities. 

\subsubsection{The back-door criterion}
To address this issue, we employ Pearl's \emph{back-door} criterion \cite{pearl2009causality}, which consists in identifying a set of adjustment variables $\mathbf{Z}$ satisfying the following two conditions:

\begin{itemize}
    \item[-] No variable in $\mathbf{Z}$ is a descendant of $X$;
    \item[-] The set $\mathbf{Z}$ blocks every back-door path between $X$ and $Y$.
\end{itemize}
Unlike directed paths that represent the flow of causal influence, a back-door path is any non-causal path between $X$ and $Y$ that begins with an arrow pointing into $X$. Such paths introduce spurious non-causal correlations between the two variables. The objective of applying the criterion is therefore to identify a set of observed variables $\mathbf{Z}$ such that, when conditioned on, the causal effect of $X$ on $Y$ can be estimated from observational data. Conceptually, $\mathbf{Z}$ acts as a filter that blocks all back-door paths, that is, all non-causal pathways through which spurious associations between $X$ and $Y$ may arise  ---thereby emulating the conditions of a randomised controlled trial.
If such a set $\textbf{Z}$ exists, our post-intervention distribution of interest can thus be recovered from purely observational probability distributions using the following adjustment formula \cite{pearl2009causality}:
\begin{equation}
P(Y \mid do(X)) = \sum_{z} P(Y \mid X, \textbf{Z}) P(\textbf{Z})
\label{Eq:general_backdoor_formula}
\end{equation}
The formula effectively re-weights the observational data distribution $P(Y \mid X, \textbf{Z})$ to remove the bias introduced by the variables in $\textbf{Z}$. 

\subsubsection{d-separation}
\label{section:d-separation}

To formally determine whether a set $\mathbf{Z}$ `blocks' all back-door paths in our causal graph $\mathcal{G}$ (shown in Figure \ref{fig:final_dag}), we employ the rules of $d$-separation (directional separation) \cite{geiger1990d}. These rules define how information flows through the three fundamental types of patterns in a causal network:
\begin{itemize}
\item[-] Chains ($X \rightarrow Z \rightarrow Y$; $X \leftarrow Z \leftarrow Y$) and forks ($X \leftarrow Z \rightarrow Y$): information flows between $X$ and $Y$ unless the intermediate variable $Z$ is conditioned on.
\item[-] Colliders ($X \rightarrow Z \leftarrow Y$): information is naturally blocked at $Z$. However, conditioning on $Z$ (or its descendants) `opens' the path, creating a spurious association between $X$ and $Y$.
\end{itemize}
To find a suitable adjustment set, we therefore examine the mutilated graph, denoted as $\mathcal{G}_{\underline{X}}$, which is the original graph $\mathcal{G}$ where we remove all edges directed out of the treatment $X$. In this sub-graph, any surviving path between $X$ and $Y$ is a back-door path. A set $\mathbf{Z}$ is thus a sufficient adjustment set if it $d$-separates $X$ and $Y$ within $\mathcal{G}_{\underline{X}}$, that is, if it blocks all chains and forks while avoiding conditioning on colliders, in accordance with the rules outlined above.

\subsubsection{Treatment effect of parking provision ($V_7$)}
\label{sec:treatment_effect_parking_prov}
To answer our primary research question, namely, assessing the causal effect of parking provision ($V_7$) on EV ownership status and intentions ($Y$) we identified $\textbf{Z}= \{V_8, V_9 \}$ to be a sufficient adjustment set to recover the post-treatment distribution $P( Y | do(V_7))$, thus yielding the following adjustment formula:  
\begin{equation}
P(Y \mid do(V_7)) = \sum_{v_8, v_9} P(Y \mid V_7, V_8, V_9) P(V_8, V_9)
\label{Eq:backdoor_formula_Parking_provision}
\end{equation}

\begin{figure}[ht!]
\centering
  \includegraphics[width=16.5cm]{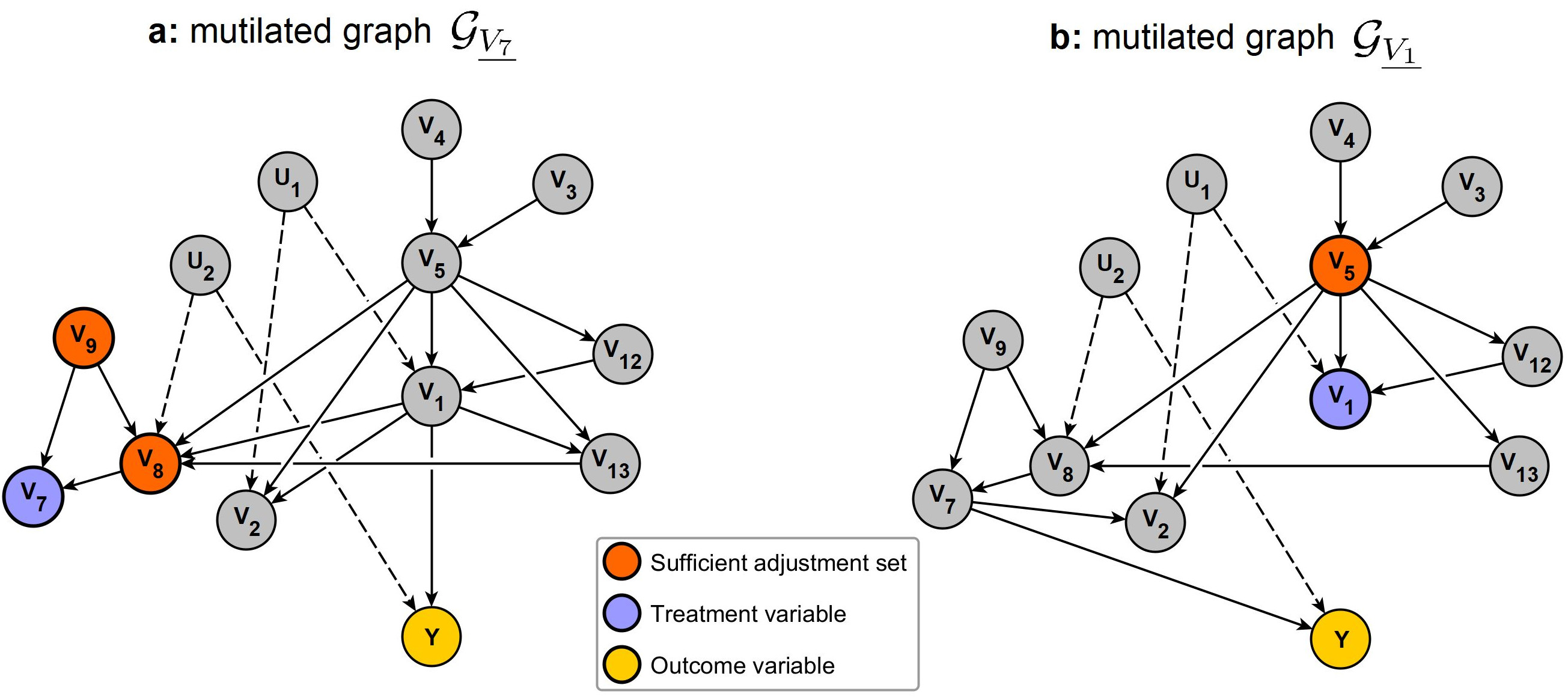}
  \caption{Mutilated graphs used for the identification of causal effects via the back-door criterion. (a) Sub-graph $\mathcal{G}_{\underline{V_7}}$ used to identify the effect of parking provision ($V_7$) on EV intentions, $Y$. (b) Sub-graph $\mathcal{G}_{\underline{V_1}}$ used to identify the effect of household income ($V_1$) on $Y$.}
  \label{fig:mutilated_graphs}
\end{figure}

\begin{proof}
Following the rules of $d$-separation, we identify a total of 24 simple\footnote{A simple path is one in which no node is visited more than once.} back-door paths between $V_7$ and $Y$ in the mutilated graph $\mathcal{G}_{\underline{V_7}}$, with this latter shown in Figure \ref{fig:mutilated_graphs}-a. The 24 paths are reported in \ref{sec:backdoorPaths_in_V7}. 
Crucially, every one of these paths is comprised of one of two distinct topological segments located immediately adjacent to the treatment $V_7$:

\begin{itemize}
\item[-] Chain segments (12 paths): these follow the pattern $V_7 \leftarrow V_8 \leftarrow \dots Y$. In these instances, $V_8$ acts as a non-collider intermediate variable (a mediator). Hence, conditioning on $V_8$ is necessary and sufficient to block the flow of non-causal information along these specific segments.
\item[-] Collider segments (12 paths): these involve a junction where $V_8$ acts as a collider: $V_7 \leftarrow V_9 \rightarrow V_8 \leftarrow \dots Y$. These segments are naturally blocked in the observational distribution by the collider at $V_8$.
\end{itemize}

While controlling for $V_8$ successfully blocks the first class of back-door segments, it simultaneously `opens' the second class due to collider bias. To resolve this, we include \emph{Dwelling age} ($V_9$) in the conditioning set. Since $V_9$ is a non-descendant of the treatment and acts as a fork (common cause) in this second class of segments, its inclusion re-establishes $d$-separation. Consequently, the set $\mathbf{Z} = \{V_8, V_9\}$ is sufficient to block all 24 back-door paths, satisfying the independence condition:
\begin{equation}
V_7 \perp Y \mid \{V_8, V_9\} \ \  \text{in} \ \ \mathcal{G}_{\underline{V_7}}
\end{equation}
This confirms that the back-door criterion is met for identification.
\end{proof}

Notably, an initial attempt to establish causal identifiability involved adjusting solely for \emph{Household income}, $\mathbf{Z}=\{V_1\}$. As visualised in the mutilated graph $\mathcal{G}_{\underline{V_7}}$ (Figure \ref{fig:mutilated_graphs}-a), conditioning on $V_1$ successfully blocks the majority of back-door paths. However, it fails to intercept the non-causal association flowing through the latent confounder \emph{Household wealth \& assets} ($U_2$), specifically along the path:
\begin{equation}
V_7 \leftarrow V_8 \leftarrow U_2 \rightarrow Y
\end{equation}
Indeed, such path could be blocked by conditioning on the common-cause variable $U_2$ if we had observational data for it. 

Beyond graphical identifiability, computation of Eq.~(\ref{Eq:backdoor_formula_Parking_provision}) requires the positivity assumption to hold, namely, that both parking provision states occur with non-zero probability within each stratum defined by the adjustment variables $V_8$ and $V_9$, i.e., $P(V_7 = v_7 \mid V_8= v_8, V_9= v_9) > 0$. We assessed this assumption empirically by examining the joint cell counts of parking provision across all 42 strata ($\vert{}V_8\vert{} \times \vert{}V_9\vert{}$) defined by dwelling type and dwelling age
\textcolor{Black}{in both the raw (unweighted) sample and the weighted sample dataset used for estimation of the network parameters (section \ref{sec:Population_weighting_resampling}).}
While most strata contain both parking categories, some exhibit separation and others show strong imbalance, thus indicating limited overlap in some parts of the covariate space. 
\textcolor{Black}{In the unweighted sample,} 4 strata (comprising 140 households, or 7.8\% of the sample) lack dual representation for both parking categories $V_7 = \text{Off-street}$ and $V_7 = \text{On-street}$, thus violating empirical positivity.
\textcolor{Black}{In the weighted sample pseudo-population, 7 strata (comprising 255 households, or 14.2\% of the sample) lack dual representation, as random weighted resampling across sparse strata introduces additional zero-frequency events in under-represented cells.}

This partial overlap reflects structural dependencies in the data-generating process, whereby parking provision is strongly associated with dwelling characteristics (e.g., certain dwelling types predominantly provide either off-street or on-street parking).
In the Bayesian network implementation, Eq.~(\ref{Eq:backdoor_formula_Parking_provision}) remains computationally well-defined despite these empirical violations because the network parameters (CPTs) are estimated introducing a small Laplace smoothing term (Eq. (\ref{eq:MLE})).
This corresponds to placing a symmetric Dirichlet prior over the conditional distributions, implying the assumption that all configurations are, a priori, equally likely. Such a weakly informative prior acts as a conservative regularisation, enforcing strictly positive probability mass on all conditional parameter configurations; consequently, zero-frequency events in the data are regularised to have non-zero (albeit very small) probability in the fitted model, ensuring a fully specified joint distribution over ($V_7, V_8, V_9$).
As mentioned, this regularisation is to be interpreted as a modelling assumption about distributional smoothness rather than empirical evidence. As a result, while estimation remains computable within the adopted Bayesian network framework, inference in those low-overlap strata relies more heavily on the structural (graphical) assumptions encoded in the model (and the choice of prior) and less on direct within-stratum empirical comparisons.

\subsubsection{Treatment effect of Household income ($V_1$)}
To answer our secondary research question, namely, assessing the causal effect of household income ($V_1$) on EV ownership status and intentions ($Y$) we identified $\textbf{Z}= \{V_5 \}$ as a sufficient adjustment set to recover the post-treatment distribution $P( Y | do(V_1))$, thus yielding to the following adjustment formula: 
\begin{equation}
P(Y \mid do(V_1)) = \sum_{v_5}{P\left(Y\mid V_1,V_5\right) P\left(V_5\right)}
\label{Eq:backdoor_formula_income}
\end{equation}

\begin{proof}
We identify, via \emph{d}-separation algorithm, a total of 27 undirected back-door paths between $V_1$ and $Y$ in the mutilated graph $\mathcal{G}_{\underline{V_1}}$, shown in Figure \ref{fig:mutilated_graphs}-b. 
These paths (reported in \ref{sec:backdoorPaths_in_V1}) can be grouped into three distinct topological classes based on their initial segments adjacent to the treatment node $V_1$:
\begin{itemize}
\item [-] Chain segments (9 paths): these paths follow the pattern $V_1 \leftarrow V_{12} \leftarrow V_5 \rightarrow \dots Y$. Here, $V_5$ acts as an ancestor to the treatment through  variable $V_{12}$ (\emph{Household working status}).
\item [-] Fork segments (9 paths): these follow the pattern $V_1 \leftarrow V_5 \rightarrow \dots Y$, where $V_5$ (\emph{Household composition}) is a direct common parent of both the treatment and subsequent causal chains.
\item[-] Collider segments (9 paths): these include the pattern $V_1 \leftarrow U_1 \rightarrow V_2 \leftarrow \dots Y$, and as such are characterised by a v-structure at the node $V_2$ (\emph{No. of vehicles}), where the latent $U_1$, \emph{Occupational classification}, and other variables converge.
\end{itemize}
By conditioning on $V_5$, we successfully $d$-separate $V_1$ from $Y$ in the subgraph $\mathcal{G}_{\underline{V_1}}$ across the first 18 paths (first and second class above), as $V_5$ acts as a common-cause intermediate node in every instance. The remaining 9 paths (third class above) are \emph{naturally} blocked by the collider at $V_2$. Because our adjustment set $\mathbf{Z} = \{V_5\}$ does not include $V_2$ or any of its descendants, these paths remain closed. Consequently, the single variable $V_5$ is sufficient to satisfy the back-door criterion, which  implies the following conditional independence condition is verified:
\begin{equation}
V_1 \perp Y \mid \{V_5\} \ \  \text{in} \ \ \mathcal{G}_{\underline{V_1}}
\end{equation}
\end{proof}

\subsection{Observational distributions}
All of the required observational distributions appearing on the r.h.s. of Eq. (\ref{Eq:backdoor_formula_Parking_provision}) and Eq. (\ref{Eq:backdoor_formula_income}) were derived  by direct querying of the Bayesian network, including the conditional probability distribution $P(Y \mid V_7)$, shown here in Figure \ref{fig:result_a} for comparison with its post-interventional counterpart $P(Y \mid do(V_7))$.
To ensure the precision of these results, these observational queries were computed using Variable Elimination (VE), a recursive algorithm designed for inference in probabilistic graphical models \cite{zhang1994simple}. VE simplifies the computation of marginal and conditional probabilities by iteratively marginalising out non-queried variables and combining the resulting \emph{potentials} until only the target distribution remains.
For this study, the Bayesian network modelling and the VE exact inference algorithm were implemented using the \texttt{pyAgrum} Python library \cite{pmlr-v138-ducamp20a}.
Unlike approximate inference methods such as those based on Markov Chain Monte Carlo (MCMC) sampling, VE yields \emph{exact} results, meaning that the output is a closed-form solution derived directly from the network's factorised joint distribution (Eq. (\ref{Bayesian_factorisation})). As such they are  not subject to approximation errors arising from sampling variability or convergence criteria.

\section{Results}

\subsection{Causal effect of parking provision}

\begin{figure}[ht!]
\centering
  \includegraphics[width=15cm]{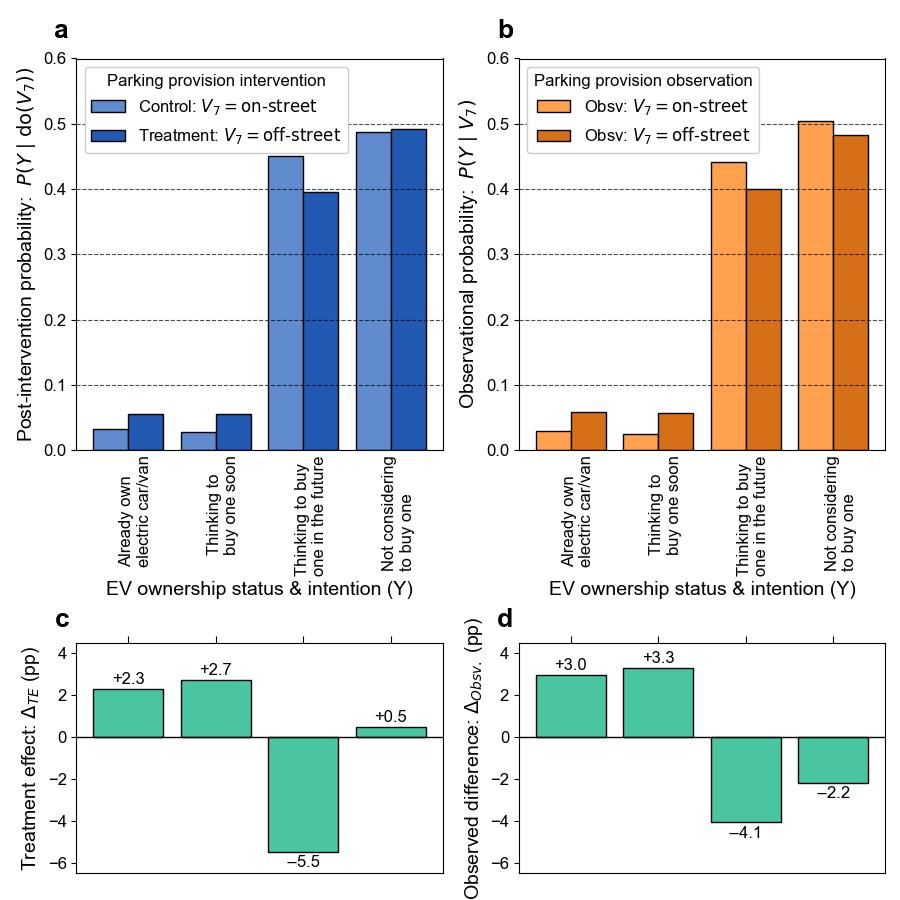}
  \caption{Probability distributions of EV adoption status and intentions ($Y$) as a result of intervening or observing parking provision ($V_7$). (a) Post-intervention probability distributions comparing households with and without off-street parking. (b) Raw observational probability distributions. (c) Causal treatment effects ($\Delta_{TE}$), illustrating the redistribution of probability mass from future intent toward immediate intent and ownership. (d) Observational shifts ($\Delta_{obs.}$), highlighting the selection bias in the ``Not considering to buy one'' category (-2.2 pp) which is absent in the causal estimation.}
  \label{fig:result_a}
\end{figure}

The causal effects of providing off-street parking across the four categories of EV ownership status and intention ($Y$) are illustrated in Figure \ref{fig:result_a}. These results provide the empirical basis for addressing RQ1 regarding the structural influence of parking provision on EV adoption. Specifically, Figure \ref{fig:result_a}-a visualises the overall post-treatment probability distribution $P(Y \mid do(V_7))$ whereas in Figure \ref{fig:result_a}-c it is shown the treatment effect $\Delta_{TE}$, measured as a contrast (difference) between the control state $do(V_7 = \text{off-street})$ and treatment state $do(V_7 = \text{on-street})$:
\begin{equation}
\label{eq:treatment_eff_PP}
    \Delta_{TE} = P(Y \mid do(V_7=\text{off-street})) - P(Y \mid do(V_7=\text{on-street}))
\end{equation}

With reference to Figure \ref{fig:result_a}-c, we can therefore observe that enabling a household to access off-street parking yields a +2.3 percentage point (pp) probability lift in the ``Already own'' category, alongside a +2.7 pp lift in immediate intent (``Thinking to buy one soon''). Conversely, this intervention produces a distinct -5.5 pp contraction in the ``Thinking to buy in the future'' category. Notably, the probability score for the ``Not considering to buy one'' category remains essentially stable, with an almost negligible shift of approximately +0.5 pp from a baseline probability of 48.7\% (i.e., $\approx1\%$ change in relative terms). We interpret this negligible shift as statistical noise inherent to the network's sensitivity rather than a meaningful causal effect (see section \ref{section:non_significant_shift}). 

Although these estimated lifts of +2.3 pp (already own) and +2.7 pp (thinking to buy soon) may appear modest in absolute terms, their magnitude is substantial when evaluated relative to the corresponding baseline probabilities: as shown in Figure \ref{fig:result_a}-a, in the absence of off-street parking, the probability of already owning an EV (in Scotland, 2022) is only 3.34\%; enabling off-street parking would increase this to 5.63\%, which corresponds to an approximate 70\% relative increase in EV ownership probability. A similar proportional amplification is observed for immediate purchase intent. Hence, while the absolute probability shifts are numerically small, they represent sizeable relative gains within categories that are initially characterised by low baseline prevalence within the population.

Importantly, these proportional increases do not arise in isolation, but correspond to a systematic reallocation of probability mass across EV adoption categories. The distribution of the treatment effects $\Delta_{TE}$ therefore indicates a higher relative concentration of probability mass in advanced EV adoption states.

Specifically, the combined +5.0 pp increase in immediate intent (2.7 pp) and ownership (2.3 pp) is largely offset by a -5.5 pp contraction in the ``Thinking to buy in the future'' category under the intervention. This pattern is consistent with a redistribution of probability mass from latent, long-term intent toward more advanced intentions of adoption, rather than a decline in overall interest.

\begin{figure}[ht!]
\centering
  \includegraphics[width=8.5cm]{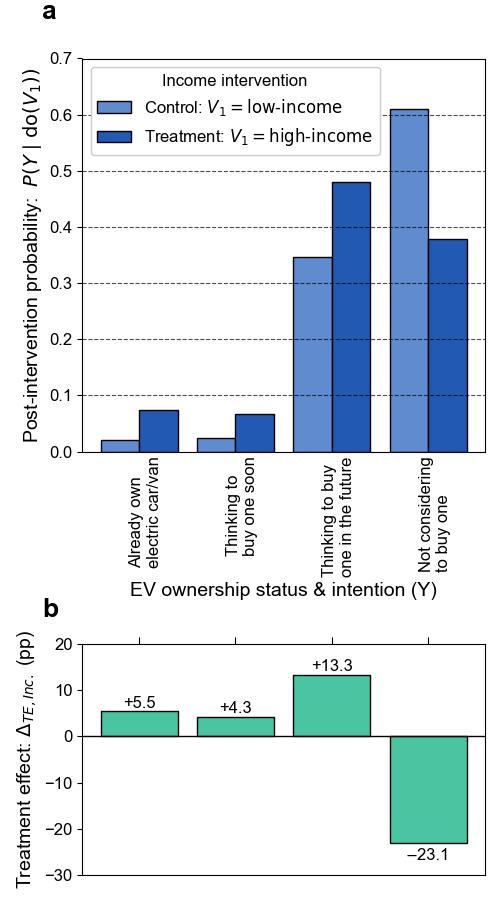}
  \caption{Total causal effect of household income ($V_1$) on EV adoption ($Y$). (a) Post-intervention distributions comparing high vs. low income. (b) Total treatment effect ($\Delta_{TE,Inc.}$) accounting for both direct affordability ($V_1 \rightarrow Y$) and indirect mediation ($V_1 \dots \rightarrow V_7 \rightarrow Y$). The -23.1 pp drop in non-participation and +13.3 pp rise in future intent indicate income as the primary driver for joining the EV adoption process.}
  \label{fig:result_b}
\end{figure}

As a result, the near-zero treatment effect for the ``Not considering to buy'' category suggests that, at the aggregate level, lack of off-street parking is unlikely to be an important constraint for households that report no intention to adopt an EV.

This distinction becomes critically apparent when comparing the causal treatment effect against the raw observational metrics, a comparison necessitated by RQ3. If, instead of estimating $\Delta_{TE}$, we had based our analysis on the observational probability shift of EV uptake conditional on parking provision:
\begin{equation}
\Delta_{obs.} =  P(Y \mid V_7=\text{off-street}) - P(Y \mid V_7=\text{on-street})
\end{equation}
we would have erroneously concluded that off-street parking access exerts an influence (however small) in pulling households away from the ``Not considering to buy'' category. As visualised in Figures \ref{fig:result_a}-b and \ref{fig:result_a}-d, the observational difference $\Delta_{obs.}$ yields a -2.2 pp reduction for this cohort. However, this is merely a spurious association arising from confounding bias, due to the influence of variables like income ($V_1$) and wealth ($U_2$) which simultaneously affect both housing choice and vehicle preference. This divergence confirms the hypothesis in RQ3; that relying on observational data alone would mislead policymakers to overestimate the impact of policies aimed at recreating the off-street charging `experience' (such as cross-pavement charging grants) on the most disengaged households while underestimating the deeper socio-economic barriers to market entry.

\subsection{Causal effect of income}
\label{sec:causal_effect_result_of_income}
While the effect invariance of the ``Not considering to buy’’ cohort to parking interventions indicates that off-street parking availability is not a major barrier, it also points toward a more fundamental constraint. As the DAG in Figure \ref{fig:final_dag} illustrates, the total causal effect of household income ($V_1$) on EV adoption ($Y$) is bipartite: first, there is a direct causal influence ($V_1 \rightarrow Y$), representing the raw financial capacity to clear the higher entry cost of an electric vehicle; second, there is an indirect influence mediated by parking provision ($V_1 \dots\rightarrow V_7 \rightarrow Y$). This latter path reflects a critical socio-economic reality: higher income is a common cause of residing in a dwelling type ($V_8$) that facilitates off-street parking ($V_7$), which in turn enables EV adoption ($Y$).
With reference to Figure \ref{fig:result_b}, by estimating the total causal effect of an income intervention as:
\begin{equation}
\label{eq:treatment_eff_income}
\Delta_{TE, Inc.} = P(Y \mid do(V_1=\text{high-inc.})) - P(Y \mid do(V_1=\text{low-inc.}))
\end{equation}
we observe a substantial -23.1 pp contraction in the probability of the ``Not considering to buy’’ category. Crucially, and in marked contrast to the parking provision results, this probability mass is mostly converted into future intent (+13.3 pp), thus signalling that income is the primary lever for shifting households from non-participation to latent interest.
\textcolor{Black}{It is important to emphasise that, consistent with the discussion on non-manipulable causes (section \ref{sec:research_gap}), an intervention on income in the causal graph should not be interpreted as a literal policy lever but rather as a structural diagnostic for the bundle of socio-economic changes that typically entail a shift in income.}

These findings highlight a clear structural hierarchy that addresses RQ2: while parking provision acts as a driver of realised EV ownership relative to latent intent to buy, income serves as the fundamental prerequisite for entering the EV adoption pipeline. This confirms that for the ``Not considering'' cohort, the primary barrier is not the so-called charging divide but an affordability ceiling that must be cleared before infrastructural interventions become relevant.

\section{Refutation tests}
\label{subsec:refutation}
Estimating causal effects relies on structural assumptions about the underlying data-generating process (see Methods section \ref{sec_methods}). Crucially, even when a causal graph is discovered from data, the resulting graph structure remains a model-based hypothesis of the invariant laws governing reality. These laws are, by definition, inaccessible; we only have access to the data they generate. This creates a distinct challenge compared to predictive machine learning. In prediction, cross-validation allows for performance evaluation against observed labels (the ground `truth'). In causal inference, such direct validation is impossible because the underlying data-generating process ---i.e., the ground-truth mechanism we aim to benchmark our model against--- is never directly observed.

To address this challenge, we follow on the work by Sharma et al. \cite{sharma2021dowhy}, which builds on the philosophical principle of \emph{falsifiability} \cite{popper1934logic}. According to this principle, scientific claims cannot be definitively proven; instead, their credibility increases when they withstand systematic attempts at refutation. Following this perspective, we implement a series of refutation tests to assess the robustness of our causal model. These tests examine whether the estimated treatment effects align with the theoretical expectation required for the model to yield valid causal conclusions. The specific expectations vary by test and are detailed in the subsections that follow.

\subsection{Placebo treatment test}
We conduct a placebo treatment test \cite{eggers2024placebo} by randomly permuting the values of the treatment variable $V_7$ within the training dataset. Specifically, each household unit is assigned a treatment value $v_7$ drawn at random from the original dataset, thereby redistributing treatment values across units. This procedure replaces the original treatment variable $V_7$ with a randomized counterpart, $V_{7,\text{placebo}}$, effectively `breaking' any causal relationship between $V_7$ and its child variables ${Y, V_2}$, while maintaining its marginal probability distribution: 
\begin{equation}
\label{eq:placebo_1}
\begin{split}
    & P(V_{7,\text{placebo}}) = P(V_7) \\
    & P(Y, V_2 \mid V_{7,\text{placebo}}) \neq P(Y, V_2 \mid V_7) \\
\end{split}
\end{equation}

We then assess the estimated causal effect of parking provision on EV uptake within this placebo framework. Under valid causal assumptions, the theoretical expectation is that the model should produce a null treatment effect, $\Delta_{TE}$, when the treatment is randomised. Although this test does not directly evaluate the estimated treatment effect, $\Delta_{TE}$, finding a significant effect where none should exist would call the model’s validity into question, and thus its estimate as well.
Formally, we state the following null hypothesis: the estimated true treatment effect of parking provision on the probability increase of owning an electric car or van ($\delta_{TE} = +2.3$ pp) is not statistically distinguishable from the effect of a random noise (placebo) treatment, $\delta_{\text{placebo}}$.

To evaluate this hypothesis, we repeat the random reassignment procedure $n$ times, generating $n = 1000$ separate placebo datasets. For each $i$-th dataset, we estimate the corresponding placebo treatment effect, $\delta_{i,\text{placebo}}$. These estimates are then used to construct a reference distribution under the null hypothesis. A $p$-value is then calculated by comparing the magnitude of $\delta_{TE}$ with the empirical distribution of the placebo effects: 
\begin{equation}
\label{eq:p-val_eq_A}
    p\text{-value}=\frac{1+\sum_{i=1}^{n}\mathbb{I}(|\delta_{i,\text{placebo}}| \geq |\delta_{TE}|)}{1+n}
\end{equation}
where $\mathbb{I}(\cdot)$ is the indicator function that counts the number of $i$ instances of $\delta_{\text{placebo}}$ that are as extreme as the estimated true effect, $\delta_{TE}$. In other words, the $p$-value represents the probability of observing a placebo treatment effect at least as large as $\delta_{TE}$ under the null hypothesis that the treatment has no causal effect on the outcome. As such, the resulting $p$-value = 0.024 (second column of Table \ref{table:refutation}) provides strong evidence against the null hypothesis. Additionally, Table \ref{table:refutation} reports the mean and median of the empirical distribution of $\delta_{\text{placebo}}$, both close to zero ($\approx-0.02$ percentage points), whereas the estimated true effect, $\delta_{TE} = +2.3$ percentage points, is substantially larger in comparison.

\subsubsection{Sensitivity analysis of non-significant probability shift}
\label{section:non_significant_shift}
While the low $p$-value ($= 0.024$) for the ``Already own electric car/van'' category validates the presence of a robust causal signal, we apply the same placebo test to the negligible shift observed in the ``Not considering to buy one'' category ($\delta_{TE}' = +0.5$ pp). In this context, we state the following null hypothesis: the estimated treatment effect of parking provision on the most disengaged cohort ($\delta_{TE}'$) is not statistically distinguishable from the effect of a random noise (placebo) treatment, $\delta_{\text{placebo}}'$. 
Unlike the previous placebo treatment test, whereby the objective was to reject the null hypothesis via a low $p$-value, now we seek to demonstrate that the estimated true effect $\delta_{TE}'$ is statistically indistinguishable from zero. Accordingly, the theoretical expectation for this negative control is a high $p$-value, indicating that the observed shift falls well within the high-density region of the null distribution. This $p$-value is computed again via Eq. (\ref{eq:p-val_eq_A}) but substituting the categorical treatment effect $\delta_{TE}$ with $\delta_{TE}'$ and the placebo effects $\delta_{i,\text{placebo}}$ with $\delta_{i, \text{placebo}}'$. 

Upon re-execution with $n = 1000$ placebo datasets, the equation yields a $p$-value = 0.817, meaning that a probability shift as extreme as $\pm0.5$ pp occurs in approximately 82\% of the randomly permuted datasets, where no causal relationship exists. Consequently, we fail to reject the null hypothesis for this category, thus supporting our interpretation that the +0.5 pp lift is due to sampling variance within the learned network parameters (CPTs) rather than the result of an invariant causal mechanism.

\begin{table}[t!]
\caption{Robustness checks for the estimated treatment effect of parking provision on the probability lift of owning an electric car or van ($\delta_{TE}$) and the lift in probability of not considering to buy one ($\delta_{TE}'$). The placebo test (columns 2–3) evaluates whether a treatment effect persists when the treatment values are randomly reassigned. The subsample test (column 4) assesses the stability of the estimate across random subsamples ($\approx$ 80\% of the full training dataset). Column 5 ($\delta_{\text{bootsr.}}$) provides the empirical bootstrap distribution used to quantify sampling uncertainty under data sparsity. Columns 6–7 report the baseline estimates of the true causal effect.}
\centering
\footnotesize
\label{table:refutation}
\addtolength{\tabcolsep}{-2.5pt}
\begin{tabular}{|l|c|c|c|c|c|c|}
\hline
& $\delta_{\text{placebo}}$ & $\delta_{\text{placebo}}'$ & $\delta_{\text{sub.}}$ & \textcolor{Black}{$\delta_{\text{bootsr.}}$} & \makecell[c]{$\delta_{TE}$ \\ (baseline)} & \makecell[c]{$\delta_{TE}'$ \\ (baseline)} \\
\hline
Mean & $-0.021$ pp & $+0.008$ pp & $+2.292$ pp & \textcolor{Black}{$+2.586$ pp} & \multirow{5}{*}{$+2.3$ pp} & \multirow{5}{*}{$+0.5$ pp} \\
Median & $-0.020$ pp & $-0.011$ pp & $+2.303$ pp & \textcolor{Black}{$+2.597$ pp} & & \\
Perc. (1st; 99th) & ($-2.48$; $+2.34$) & ($-4.98$; $+5.48$) & ($+1.20$; $+3.41$) & \textcolor{Black}{($+0.26$; $+4.77$)} & & \\
$p$-value\textsuperscript & 0.024 & 0.817 & 0.983 & — & & \\
Sample size $n$ & 1000 & 1000 & 1000 & \textcolor{Black}{1000} & & \\
\hline
\end{tabular}
\end{table}

\subsection{Data subsample test}
\label{sec:Data_sub-sample_test}
This refutation test evaluates the robustness of the estimated effect of parking provision by assessing its sensitivity to variations in the training dataset through repeated random subsampling. Specifically, we generate $n = 1000$ random subsamples from the full training dataset, each containing approximately 80\% of the household units, and compute the corresponding treatment effect estimate, $\delta_{i, \text{sub.}}$, for each iteration. 
\textcolor{Black}{As for the placebo test, $\delta_{i,\text{sub.}}$ is the probability lift of the ``already owns an electric car/van'' state.}

This procedure yields an empirical distribution of subsample-based estimates, $\delta_{\text{sub.}}$, against which the full-sample estimate, $\delta_{TE}$, is compared. 
The underlying intuition is that a stable estimate should not be overly influenced by any specific subset of the data and should, therefore, lie well within the range of the subsample-based distribution. 
Formally, we test the null hypothesis that the full-sample effect is not significantly different from the effects estimated across these random draws. In this context, a high $p$-value is desirable as it indicates that the full-sample estimate's position relative to the mean of the subsamples is consistent with random sampling variability. 
The $p$-value is calculated as:
\begin{equation}
\label{p-value_eq_b}
p\text{-value} = \frac{1+\sum_{i=1}^{n} \mathbb{I} (|\delta_{i, \text{sub.}} - \delta^{*}_{\text{sub.}} | \geq |\delta_{TE} - \delta^{*}_{\text{sub.}}| )}{1+n}
\end{equation}
where:
\begin{itemize}
\item[-] $\delta_{i, \text{sub.}}$ is the treatment effect estimated from the $i$-th subsample dataset;
\item[-] $\delta^{*}_{\text{sub.}}$ is the mean of effect estimates of the $n$ subsample datasets;
\item[-] $\delta_{TE}$ is the effect estimate obtained from the full dataset;
\item[-] $\mathbb{I}(\cdot)$ is the indicator function that counts how many subsample deviations from the mean equal or exceed the deviation of the full-sample estimate from that same mean.
\end{itemize}
Retaining the null hypothesis (evidenced by a high $p$-value, possibly close to 1) strengthens the credibility of the causal estimate by demonstrating that it is not hypersensitive to any particular partition of the training data. 
As reported in the fourth column of Table \ref{table:refutation}, the estimated full-sample effect of $\delta_{TE} = +2.3$ pp lies well within the empirical distribution and is nearly identical to the distribution's mean and median. 
The resulting $p$-value of 0.983 confirms that the full-sample estimate is not an outlier relative to the observed subsample variability; consequently, we fail to reject the null hypothesis, providing strong evidence of the estimate's stability.

A methodological limitation of this subsample-based refutation test in our setting must be noted. Because the underlying training dataset is constructed using survey weighting with replacement, drawing 80\% random subsamples results in overlap of duplicated household units across draws, artificially reducing the variance of the subsample estimates $\delta_{i,\text{sub.}}$. The results of this refutation test should therefore be interpreted cautiously as a diagnostic of internal consistency rather than true out-of-sample stability.

\subsection{Uncertainty quantification under sparse data sampling}
\label{subsec:bootstrap_uncertainty}
We further quantify uncertainty in the estimated treatment effect, $\delta_{TE}$, arising from the limited effective sample size of the training data. As discussed in Section~\ref{sec:Population_weighting_resampling}, only a small number of unique household observations are represented in certain strata, increasing sampling variability and reducing estimator precision.
To assess this uncertainty, we employ a survey-weighted non-parametric bootstrap procedure. Specifically, we generate $n = 1000$ bootstrap replicates by resampling household units with replacement from the original unweighted dataset, using selection probabilities proportional to the survey weight variable (\texttt{tsWghtP\_n}). For each bootstrap replicate, survey weights are re-applied, the Bayesian network conditional probability tables (CPTs) are re-estimated, and the causal estimand is recomputed under the refitted model. 
This approach follows the bootstrap plug-in principle, whereby the weighted empirical distribution of the observed sample serves as a proxy for the underlying population distribution. By repeatedly resampling from this empirical distribution, we approximate repeated sampling from the target population and obtain an empirical estimate of the sampling variability of the treatment effect estimator.
The resulting bootstrap estimates, denoted by $\delta_{\text{bootstr.}}$, are reported in the fifth column of Table~\ref{table:refutation}. Crucially, the baseline point estimate ($\delta_{TE} =+2.3$ pp) remains highly congruent with both the empirical bootstrap mean ($+2.59$ pp) and median ($+2.60$ pp),

It should be noted however the presence of  substantial dispersion of the empirical bootstrap estimate, with a 1st--99th percentile interval ranging from +0.26 pp to +4.77 pp, reflecting the expected loss in precision associated with sparse sampling.

In addition to the percentile bootstrap interval, we report the corresponding \emph{basic bootstrap interval} following the construction described by Davison and Hinkley~\cite{davison1997bootstrap}. For our original point estimate $\delta_{TE} = +2.3$~pp and bootstrap quantiles $(+0.26, +4.77)$, the basic bootstrap interval is given by:
\begin{equation}
\left(2\delta_{TE} - 4.77, \; 2\delta_{TE} - 0.26\right) = \left(-0.17, \; +4.34\right)
\end{equation}
The percentile and basic bootstrap intervals provide complementary characterisations of uncertainty. Because the empirical bootstrap mean ($+2.59$~pp) slightly exceeds the original point estimate ($\delta_{TE}=+2.3$~pp), the basic bootstrap applies an explicit bias-correction shift downward around $\delta_{TE}$. Consequently, while the percentile interval remains strictly positive $(+0.26, +4.77)$, the basic interval lightly crosses the null $(-0.17, +4.34)$. This divergence highlights that while the central tendency of the effect remains positive across resampling, the precision of the lower bound is sensitive to the choice of bootstrap construction.

\textcolor{Black}{\subsection{Sensitivity to graph misspecification}
\label{sec:sensitivity_to_graph_misspecification}
To address potential concerns regarding graph misspecification and the reliance on a hybrid DAG (derived via the MIIC algorithm and augmented with post-hoc domain knowledge), we evaluate the sensitivity of our causal identification strategy to alternative graph topologies.
As formally demonstrated in section \ref{sec:treatment_effect_parking_prov}, a valid adjustment has been identified as $\{V_8, V_9\}$. With reference to the mutilated graph $\mathcal{G}_{\underline{V_7}}$ shown in Figure \ref{fig:mutilated_graphs}-a, this set satisfies the backdoor criterion by simultaneously blocking all open backdoor paths propagating through $V_8$ and neutralising the introduced collider stratification bias by conditioning on $V_9$ as well. Crucially, the specific graph topology surrounding the treatment node implies that the causal estimates resulting from such adjustment set $\{V_8, V_9\}$ are invariant to a wide range of alternative graph specifications. For instance introducing additional, theoretically plausible, edges (such as $Y \rightarrow V_2$, or $U_2 \rightarrow V_2$, or $U_2 \rightarrow V_9$),\footnote{Specifically, $Y \rightarrow V_2$ represents the behavioural mechanism where EV adoption prompts households to retain an additional internal combustion engine vehicle to mitigate range anxiety, thereby increasing total vehicle count. $U_2 \rightarrow V_2$ would capture the economic effect of latent household wealth/assets on the capacity to own multiple vehicles. Lastly, $U_2 \rightarrow V_9$ would model the preference of wealthier households to sort into highly valued period properties, mirroring the structural relationship encoded between income and dwelling age.} does not alter d-separation status between $V_7$ and $Y$ given $\{V_8, V_9\}$. As such, the backdoor criterion remains satisfied, meaning the exact same estimator (Eq. (\ref{Eq:backdoor_formula_Parking_provision})) and numerical estimates hold. 
The only topological modifications capable of invalidating our current estimator would be the introduction of causal edges enabling unblocked backdoor paths bypassing the controlled covariates $V_8$ and $V_9$ altogether:
\begin{equation}
\exists V_i \in \mathbf{V} \setminus \{V_8, V_9\}
:\quad
V_7 \leftarrow V_i \rightarrow \dots Y
\end{equation}
However, an inspection of the subset $\mathbf{V} \setminus \{V_8, V_9\}$, only reveals theoretically weak mechanisms that would grant a direct causal edge into parking provision ($V_7 \leftarrow V_i$). Consequently, if a misspecified backdoor path does exist, it must propagate through an unmeasured, latent confounder ($X \leftarrow U \rightarrow Y$). We evaluate the vulnerability of our estimates to such an unmeasured confounder in the following section \ref{sec:omitted_var_bias_sensitivity}.}

\subsection{Sensitivity analysis to omitted variable bias}
\label{sec:omitted_var_bias_sensitivity}

While the placebo and subsample tests provide empirical evidence of robustness within the observed data and assumed unobserved confounders, a more fundamental issue concerns the possibility of having omitted a confounder. By definition, such a confounder cannot be tested directly, yet it may still bias causal estimates if an omitted variable affects both treatment assignment and outcomes ($V_7 \leftarrow U \rightarrow Y$). To assess this risk, we conduct a sensitivity analysis that examines how strong such a confounder would need to be to fully explain away (i.e., eliminate) the estimated treatment effect $\Delta_{TE}$, Eq. (\ref{eq:treatment_eff_PP}). Although the presence of such a confounder is purely hypothetical and ultimately unknowable (potentially representing one, multiple, or no factor at all), for the sake of argument we might posit that \emph{Environmentalism} ($U$) functions as a latent confounder, simultaneously increasing the likelihood of EV ownership and the propensity to reside in more rural areas, where off-street parking is structurally more common. 
To note: this skeptical line of reasoning runs counter to the (arguably more convincing) view that environmentally conscious individuals choose dense urban areas in order to minimise their ecological footprint. If this is the case, the \emph{true} causal effect of parking provision on EV ownership would be even larger than our current estimate $\Delta_{TE}$. 

We now express the identified causal effect of parking provision on EV ownership as a risk ratio ($RR$), defined in terms of post-intervention probabilities:

\begin{equation}
\label{eq:RR_parking}
    RR_{Parking}=\frac{P(Y=\text{already-own} \mid do( V_7=\text{off-street}))}{P(Y=\text{already-own} \mid do( V_7=\text{on-street}))}=\frac{5.63\%}{3.34\%}=1.685
\end{equation}
To test this $\approx +70\%$ causal effect increase we use the $E$-value metric introduced by VanderWeele \& Ding \cite{vanderweele2017sensitivity}. This approach is particularly powerful because it is fully non-parametric, hence making no assumptions about the functional form of the association between $U$ and the treatment (and outcome) variables. The $E$-value is defined as the minimum strength of association an unmeasured confounder must have with \emph{both} the treatment and the outcome, on the risk ratio scale, to explain away the $RR$ as a null effect:

\begin{equation}
    E\text{-value} = RR + \sqrt{RR(RR-1)}
\end{equation}
By plugging in the above formula the $RR_{Parking}$ value obtained via Eq. (\ref{eq:RR_parking}) we obtain an $E$-value = 2.76 for our point estimate. This indicates that for a hypothetical unmeasured confounder like \emph{Environmentalism} to fully explain away our observed association, it would need to be associated with both parking provision and EV ownership by a risk ratio of at least 2.76-fold each. 

Following VanderWeele \& Ding \cite{vanderweele2017sensitivity} to compute the E-value also for the limit of the confidence interval closest to the null, we evaluate the statistical precision of our estimate under the sparse sampling conditions detailed in section~\ref{subsec:bootstrap_uncertainty}. Using our  bootstrap replicates, the $1\text{st}$ percentile lower limit of our Risk Ratio ($LL_{1\%}$) is estimated at 1.07. Plugging this highly conservative lower bound into the E-value formula yields an $E$-value for the confidence interval limit of 1.34, meaning that an unmeasured confounder associated with both parking provision and EV ownership by a joint Risk Ratio of at least 1.34-fold each could shift our $99\%$ confidence interval to include the null.

To evaluate the plausibility of these magnitudes, we benchmark them against the estimated causal effect of income, the primary socio-economic driver in our model, expressed this time on the risk ratio scale:
\begin{equation}
\label{eq:RR_income}
    RR_{Income}=\frac{P(Y=\text{already-own} \mid do( V_1=\text{high-inc.}))}{P(Y=\text{already-own} \mid do( V_1=\text{low-inc.}))}=\frac{7.47\%}{2.00\%}=3.73
\end{equation}

While the required bias to shift the point estimate (2.76) is numerically lower than the effect of income on EV ownership (3.73), it still represents a prohibitive threshold for any latent factor (such as environmentalism) to overcome. For a hypothetical confounder to nullify the observed point estimate effect of parking provision, it would need to exert a simultaneous causal influence on both EV ownership and parking provision equivalent to nearly three-quarters of a household's financial capacity.
Even under extreme uncertainty bounds, explaining away our conservative $1\text{st.}$ percentile lower limit still requires an unmeasured confounder with a 1.34-fold risk ratio association with both treatment and outcome.
\textcolor{Black}{This implies that a relatively modest unobserved confounder could shift the lower bound across the null. Although the failure of the graph discovery algorithm to detect a latent signature ($V_7 \leftrightarrow Y$) supports the structural independence of $V_7 \rightarrow Y$, the lower-bound $E$-value highlights that the statistical interval remains moderately sensitive to potential omitted variable bias.}

\section{Conclusions}

The transition to electric mobility represents a critical pathway toward achieving net-zero emissions, yet its success hinges on equitable access rather than mere aggregate adoption. This study addressed a fundamental gap in the existing literature: determining whether the stark ‘charging divide’ observed between households with and without off-street parking is primarily a function of physical infrastructure constraints or a by-product of underlying socio-economic disparities.

Methodologically, this research represents a novel departure from traditional transport modelling. By framing the problem within a probabilistic causal framework and applying it to a nationally representative dataset of Scottish households, we moved beyond predictive associations to quantify the interventional effect of residential parking provision. This approach allows for a level of analytical rigour that traditional regression analysis cannot achieve: the ability to simulate policy interventions directly from observational data while explicitly neutralising the influence of confounding variables (whether latent or observed).

Our causal analysis directly addressed the three primary research questions outlined in this study through a clear structural hierarchy:

\begin{itemize}
    \item[-] The conversion driver (RQ1): we estimated that the absence of private off-street parking imposes a distinct, quantifiable `adoption penalty' on households. By controlling for back-door paths, the model confirms that parking access is not merely correlated with EV uptake, but acts as a direct causal factor. Specifically, enabling off-street parking yields a 70\% relative increase in the probability of EV ownership (moving from 3.3\% to 5.6\%), corresponding to a 2.3 percentage point increase. The (1st; 99th) percentile bootstrap interval is (+0.26; +4.77) percentage points, while the corresponding basic bootstrap interval is ($-0.17$; $+4.34$) percentage points. Although the point estimate remains positive, the inclusion of zero in the basic bootstrap interval indicates that the direction of the estimated effect is sensitive to the bootstrap interval construction.
    This effect is largely a conversion mechanism; the +5.0 pp gain in immediate intent and ownership is almost entirely offset by a -5.5 pp contraction in future intent to buy an EV, suggesting that off-street parking access facilitates those already considering to purchase an EV rather than recruiting new participants.

    \item[-] The affordability ceiling (RQ2): while parking provision facilitates the final stages of EV adoption, our findings indicate that household income serves as the fundamental gatekeeper for entering the EV market. A simulated intervention on income produced a substantial -23.1 pp drop in the ``Not considering to buy [an EV]'' category, with the majority of that mass shifting into future intent to buy (+13.3 pp). This identifies a clear affordability ceiling: for the nearly 50\% of the Scottish population not engaged with the EV market, the primary barrier is financial capacity rather than the immediate lack of an off-street parking spot.

    \item[-] The necessity of causal inference (RQ3): our comparison between observational probabilities and interventional probabilities highlights the risks of relying on purely predictive (and often black-box) algorithms for decision making. Machine learning models trained with the sole purpose of predicting observational outcomes would erroneously suggest that investing in on-street home-charging infrastructure would reduce non-participation in the EV market. Our model reveals this is a statistical artifact of confounding bias; wealthy households are simply more likely to possess both the income to consider an EV and the dwelling type to park it. Mistaking this correlation for a causal effect risks leading policymakers to over-invest in infrastructure for cohorts who have not yet cleared the financial hurdles of entering the EV market.
\end{itemize}

These findings suggest that a one-size-fits-all approach to EV promotion may be inefficient. Instead, Scottish policy could reflect the different structural barriers faced by households at different stages of the EV adoption `journey'.
For households currently outside the market, the results suggest that affordability remains the primary bottleneck. Until the high entry cost of EVs is addressed (perhaps through broader secondary market support or targeted financial incentives) infrastructural interventions like on-street charging grants may see limited engagement from this cohort.
Conversely, for the `latent intent' group (those who intend to buy but are stalled in the `future' category) the charging divide is the definitive barrier. In high-density urban environments like Glasgow or Edinburgh, where tenements currently preclude domestic charging, qualitative policy shifts toward expanding public charging density and cross-pavement solutions are essential. 

By recognising income as the gatekeeper and parking as a driving factor, policymakers can design a more nuanced, equitable transition that addresses both the financial prerequisites and the physical constraints of the Scottish dwelling population.

\textcolor{Black}{In terms of study limitations, we acknowledge that the limited sample size and uneven distribution of EV-owning households across socio-economic strata introduce sampling uncertainty in parts of the covariate space. Additionally, the integration and cross-linking of multiple data sources resulted in a reduced effective sample size, although this uncertainty under sparse data sampling has been quantified via bootstrapping. Finally, results are constrained by the available variables and their measurement granularity, particularly the aggregation of charging infrastructure at the local authority level. As such, while providing policy-relevant insights into EV adoption dynamics, findings should be interpreted as conditional on the observed data structure and modelling assumptions.}

\section{Data and code availability}
\label{sec_DC_aval}
All data and code developed for this study are publicly available and can be accessed via the referenced GitHub repository \cite{github_repoEV_uptake}. 

\section{Acknowledgements}
The authors express their gratitude to the School of Computing, Engineering \& the Built Environment (SCEBE) at Edinburgh Napier University for providing the support and resources necessary for the completion of this research. 

\bibliographystyle{elsarticle-num} 
\bibliography{refs_file}

%\newpage

\appendix

\section{}

\label{sec_appendix_A}
\setlength{\LTcapwidth}{\textwidth}
% --- Add these lines to get "Table A1" style ---
\setcounter{table}{0}                % Reset table counter
\renewcommand{\thetable}{A\arabic{table}} % Prefix with 'A'
\renewcommand{\theHtable}{AppTable.\thetable}
\setcounter{figure}{0}  
\renewcommand{\theHfigure}{AppFigure.\thefigure}

\begin{table}[ht]
    \centering
    \begin{threeparttable}
    \caption{Mapping between the 9-fold classification for the \emph{Parking Provision} variable, $V_7$, (as reported in the SHCS dataset) and the corresponding binary classification $\{\text{Off-street}; \text{On-street}\}$. \label{tab:mapping_parking_provision}}
    \begin{tabular}{|l|c|} 
    \hline
        \textbf{9-fold classification} & \textbf{2-fold classification} \\
    \hline
        Integral/attached garage   & Off-street \\
        Garage on plot             & Off-street \\
        Space on plot              & Off-street \\
        Space/garage elsewhere     & Off-street \\
        Adequate, on-street        & On-street \\
        Inadequate on-street       & On-street \\
        No parking provision       & --- \\
        Not Applicable             & --- \\
        Unobtainable               & ---  \\
    \hline
    \end{tabular}
    \end{threeparttable}
\end{table}

{\scriptsize
\setlength{\LTcapwidth}{\textwidth} % Ensures caption uses full width

\begin{longtable}{|p{5cm}|p{1.5cm}|l|p{2.2cm}|}

\caption{Set $\textbf{V} = \{Y, V_1, ... ,V_{14}\}$ of observed variables used to discover and then train the model parameters of the causal Bayesian Network. \label{table:variable_list}}
 \\

\hline
\textbf{Variable label} & \textbf{Variable symbol} & \textbf{Value states} & \textbf{Probability distribution} $P(\cdot)$\\
\hline
\endfirsthead

\multicolumn{3}{l}{\textit{Table \thetable\ (continued)}} \\
\hline
\textbf{Variable name} & \textbf{Variable symbol} & \textbf{Value states}  & \textbf{Probability distribution} $P(\cdot)$\\
\hline
\endhead
\hline
EV ownership status \& intention & $Y$ & \makecell[l]{Already own electric car/van; \\ Thinking to buy one soon; \\ Thinking to buy one in the future; \\ Not considering to buy one;} & \makecell[l]{0.0485 \\ 0.0462 \\ 0.4149 \\ 0.4904} \\
\hline
Household income & $V_1$ & \makecell[l]{< £6000; \\ £6001-10,000; \\ £10,001-15,000; \\ £15,001-20,000; \\ £20,001-25,000; \\ £25,001-30,000; \\ £30,001-40,000; \\ > £40,000;} & \makecell[l]{0.0202 \\ 0.0364 \\ 0.0786 \\ 0.1358 \\ 0.1079 \\ 0.1064 \\ 0.1602 \\ 0.3546} \\
\hline
No. of vehicles & $V_2$ & \makecell[l]{No car; \\ One car; \\ Two cars or more;} & \makecell[l]{0.0719 \\ 0.6157 \\ 0.3124} \\
\hline
No. of adults in household & $V_3$ & \makecell[l]{1; \\ 2; \\ 3; \\ 4; \\ $\ge$ 5;} & \makecell[l]{0.3668 \\ 0.5400 \\ 0.0705 \\ 0.0211 \\ 0.0017} \\
\hline
No. of children in household & $V_4$ & \makecell[l]{0; \\ 1; \\ 2; \\ 3; \\ $\ge$ 4;} & \makecell[l]{0.7764 \\ 0.1149 \\ 0.0871 \\ 0.0161 \\ 0.0055} \\
\hline
Household composition & $V_5$ & \makecell[l]{Single adult; \\ Small - multiple adults; \\ Single parent; \\ Small - family; \\ Large - family; \\ Large - multiple adults; \\ Small - old adults; \\ Single pensioner;} & \makecell[l]{0.1759 \\ 0.2149 \\ 0.0901 \\ 0.1029 \\ 0.0289 \\ 0.0615 \\ 0.2030 \\ 0.1227} \\ 
\hline
Urban/rural classification & $V_6$ & \makecell[l]{Urban; \\ Rural;} & \makecell[l]{0.8069 \\ 0.1931} \\
\hline
Parking provision & $V_7$ & \makecell[l]{Off-street; \\ On-street;} & \makecell[l]{0.6516 \\ 0.3484}  \\
\hline
Dwelling type & $V_8$ & \makecell[l]{Detached house; \\ Semi-detached house; \\ Terraced house; \\ Tenement flat; \\ 4-in-a-block flat; \\ Tower/slab flat; \\ Flat from converted house;} & \makecell[l]{0.2573 \\ 0.2276 \\ 0.2073 \\ 0.1657 \\ 0.0765 \\ 0.0327 \\ 0.0329} \\
\hline
Dwelling age & $V_9$ & \makecell[l]{< 1919; \\ 1919-1944; \\ 1945-1964; \\ 1965-1982; \\ 1983-2002; \\ > 2002;} & \makecell[l]{0.2114 \\ 0.1093 \\ 0.1615 \\ 0.2214 \\ 0.1493 \\ 0.1471} \\ 
\hline
\makecell[l]{Workplace charging \\ infrastructure density \\ (normalised per 100,000 \\ residents)} & $V_{10}$ & \makecell[l]{< 20; \\ 21-40; \\ 41-60; \\ 61-80; \\ > 80;} & \makecell[l]{0.0821 \\ 0.3346 \\ 0.4562 \\ 0.0910 \\ 0.0361} \\ 
\hline
\makecell[l]{Public charging \\ infrastructure density \\ (normalised per 100,000 \\ residents)} & $V_{11}$ & \makecell[l]{< 50; \\ 51-100; \\ 101-150; \\ 151-200; \\ > 200;} & \makecell[l]{0.2048 \\ 0.5949 \\ 0.1537 \\ 0.0438 \\ 0.0028} \\
\hline
Household working status & $V_{12}$ & \makecell[l]{One or more working adults; \\ None working;} & \makecell[l]{0.6664 \\ 0.3336} \\
\hline
Tenancy & $V_{13}$ & \makecell[l]{Owned (outright or mortgage); \\ Part mortgage, part rent; \\  Rented (LA, Co-op, private landlord;)} & \makecell[l]{0.7354 \\ 0.0041 \\ 0.2605} \\
\hline
Local Authority (LA) & $V_{14}$ & \makecell[l]{South Ayrshire; \\ South Lanarkshire; \\ Stirling; \\ West Dunbartonshire; \\ West Lothian; \\ Na h-Eileanan Siar; \\ Aberdeen City; \\ Aberdeenshire; \\ Angus; \\ Argyll and Bute; \\ Scottish Borders; \\ Clackmannanshire; \\ Dumfries and Galloway; \\ Dundee City; \\ East Ayrshire; \\ East Dunbartonshire; \\ East Lothian; \\ East Renfrewshire; \\ City of Edinburgh; \\ Falkirk; \\ Fife; \\ Glasgow City; \\ Highland; \\ Inverclyde;  \\ Midlothian; \\ Moray; \\ North Ayrshire; \\  North Lanarkshire; \\ Orkney Islands; \\ Perth and Kinross; \\ Renfrewshire;  \\ Shetland Islands; } & \makecell[l]{0.0266 \\ 0.0638 \\ 0.0178 \\ 0.0205 \\ 0.0383 \\ 0.0078 \\ 0.0466 \\ 0.0433 \\ 0.0216 \\ 0.0211 \\ 0.0228 \\ 0.0072 \\ 0.0266 \\ 0.0244 \\ 0.0250 \\ 0.0183 \\ 0.0182 \\ 0.0117 \\ 0.1077 \\ 0.0327 \\ 0.0527 \\ 0.1038 \\ 0.0411 \\ 0.0128 \\ 0.0172 \\ 0.0183 \\ 0.0178 \\ 0.0594 \\ 0.0028 \\ 0.0333 \\ 0.0316 \\ 0.0072} \\  
\hline
\end{longtable}
}

%\newpage
\section{}
\label{sec_appendix_B}
This appendix provides a brief technical note on the information-theoretic causal discovery algorithm introduced in section \ref{section:CausalStructureLearning}. For a comprehensive description of the methodology, the reader is referred to the original works of Verny et al. \cite{verny2017learning}, who introduced the MIIC framework, and Ribeiro-Dantas et al. \cite{ribeiro2024learning}, who further developed and applied it to large-scale datasets.

\subsection{Information-theoretic-driven causal discovery}
\label{section:CausalMIIC}
The MIIC framework identifies the skeleton of the causal graph by systematically evaluating the information flow between variable pairs using information-theoretic metrics.
Mutual information (MI) is a non-negative measure of association (e.g. in units such as bits) shared between two random variables $X$ and $Y$:
\begin{equation}
    I(X;Y)
    = \sum_{x,y} P(x, y)
    \log\!\left[
    \frac{P(x, y)}{P(x)\,P(y)}
    \right]
    \label{Eq:MI_X_Y}
\end{equation}
where the ratio inside the logarithm compares the joint probability of the event $\{X = x,Y = y\}$ to the product of the corresponding marginal probabilities. If $X$ and $Y$ are independent, this ratio equals $1$, and hence the logarithm evaluates to $0$. The double summation in Eq. (\ref{Eq:MI_X_Y}) aggregates the information gain across all possible states (values) of $X$ and $Y$. Unlike statistical independence tests, which provide a binary ($\{\text{true}, \text{false}\}$) assessment of dependence, MI provides a continuous, non-negative measure of association strength $\in[0,\infty)$.
Similarly, conditional MI measures the amount of information shared between $X$ and $Y$ given that a third variable $Z$ is being observed:
\begin{equation}
    I(X;Y \mid Z)
    = \sum_{x,y,z} P(x, y, z)
    \log\!\left[
    \frac{P(x, y \mid z)}{P(x \mid z)\,P(y \mid z)}
    \right]
\end{equation}
If $X$ and $Y$ are conditionally independent given $Z$, then $I(X;Y \mid Z) \approx 0$, reflecting the absence of any shared information between $X$ and $Y$ once the influence of $Z$ has been accounted for, meaning, any dependence between $X$ and $Y$ is entirely explained away by their mutual relationship with $Z$. Like for mutual information, conditional mutual information is a continuous, non-negative measure.

\subsection{Iterative skeleton pruning}
As for any constraint-based causal discovery algorithm, MIIC starts with a fully connected graph, representing the initial hypothesis that every pair of variables might be dependent upon each other (see Figure \ref{fig:Causal_Discovery_workflow}-a). The algorithm then evaluates the conditional MI for every variable pair to determine if their association is direct or mediated by other variables.

\begin{figure}[ht!]
\centering
  \includegraphics[width=15.5cm]{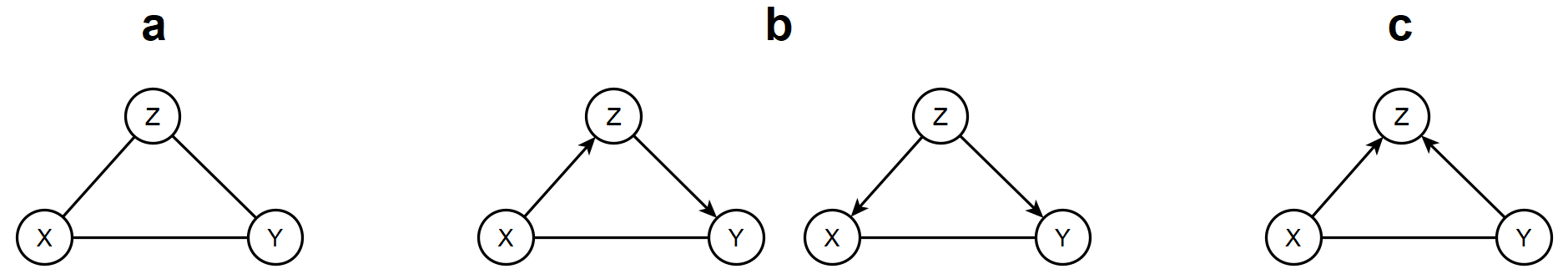}
  \caption{Information-theoretic signatures of causality in a three-node skeleton. (a) Initial skeleton: the MIIC algorithm starts with an undirected graph where $I(X;Y)$ represents the total association between $X$ and $Y$.
(b) Interaction information $I(X;Y;Z) > 0$ is a signature of a causal chain or common cause. In these patterns, $Z$ explains part of the mutual information, reducing the residual conditional mutual information between $X$ and $Y$.
(c) Interaction information $I(X;Y;Z) < 0$ is a signature of a collider (v-structure). Conditioning on $Z$ increases the conditional mutual information between $X$ and $Y$, leading to its rejection from the conditioning set $\{Z_i\}$ during skeleton pruning.}
  \label{fig:3_node_skeleton}
\end{figure}

Unlike traditional methods that may test all possible combinations of variables to condition on (an approach that becomes computationally unfeasible as the network grows) MIIC employs a greedy, iterative decomposition strategy \cite{ribeiro2024learning}: for each pair $(X, Y)$, the algorithm searches for a set of contributors $\{Z_i\}$ that explain away the shared information between them. Crucially, a contributor $Z_i$ is only accepted if the value of the \emph{interaction} information is positive, thus signalling the signature of a chain (e.g., $X\rightarrow Z_i \rightarrow Y$) or common cause ($X \leftarrow Z_i \rightarrow Y$). To build intuition, let us consider the simplest case of a three-node fully connected skeleton as shown in Figure \ref{fig:3_node_skeleton}-a. Here, we can express the MI between $X$ and $Y$ as the sum of a conditional MI contribution and an \emph{interaction} MI contribution:
\begin{equation}
\label{eq:three_node_skeleton}
    I(X;Y)
    = I(X;Y \mid Z) + I(X;Y;Z)
\end{equation}
noting that $I(X;Y \mid Z)$ measures the information `flow' shared via the direct edge $X-Y$ whereas $I(X;Y;Z)$ accounts for the association between $X$ and $Y$ flowing across the indirect path $X-Z-Y$. As per Eq. (\ref{eq:three_node_skeleton}), when the interaction information term $I(X; Y; Z)$ is negative, it follows that the conditional information $I(X;Y \mid Z)$ is larger than the overall MI $I(X;Y)$. The only case when conditioning on a third variable $Z$ increases the direct flow of association, via the $X-Y$ edge, it is when both $X$ and $Y$ are common-causes of $Z$, namely, $Z$ is a collider: $X\rightarrow Z \leftarrow Y$ (Figure \ref{fig:3_node_skeleton}-c). Since this step aims to identify variables that explain away (reduce) the association between $X$ and $Y$, colliders are rejected from the conditioning set $\{Z_i\}$ to avoid introducing biased dependencies. 

In more realistic networks involving a large number of variables, the algorithms generalises this principle, eventually isolating the unbiased direct association $I(X; Y \mid \{Z_i\})$ for every $X-Y$ edge in the network.

Having quantified the unbiased association strength for every edge, to determine whether this remaining associations represent genuine causal relations or mere sampling noise, their strength is measured against a dynamic complexity threshold: an edge is retained only if its conditional MI exceeds the cost of adding that specific dependency to the model, namely, if $I(X;Y \mid \{Z_i\}) > \mathcal{C}_{NML}(X,Y \mid \{Z_i\}) / N$.
The term $\mathcal{C}_{NML}$ denotes the Normalised Maximum Likelihood complexity \cite{Rissanen1996}, which quantifies a model's inherent capacity to fit many data patterns.

\subsection{Edge orientation and latent variables identification}
\label{subsection:edge_orient}
After obtaining the pruned skeleton, the next step entails determining causal flow via edges orientation. All unshielded node triples $\langle X; Z; Y \rangle$ (sets of nodes where $X$ and $Y$ are both adjacent to $Z$ but are not adjacent to each other) are evaluated to identify the presence of a v-structure. Because in those triples $X$ and $Y$ are non-adjacent, $Z$ must act as a mediator, a common cause, or a collider. For each triple, the correct orientation is established based on \emph{interaction} mutual information signatures. 

To build intuition, let us reconsider the 3-node example in Figure \ref{fig:3_node_skeleton}-a, assuming now the direct edge $X-Y$ was removed during pruning. If the interaction term $I(X; Y; Z) < 0$, it follows that $X$ and $Y$ are independent but become dependent upon conditioning on $Z$, implying a signature of causality for a v-structure, hence arrowheads are placed pointing toward the mid-node ($X \rightarrow Z \leftarrow Y$).
Conversely, a positive interaction term, $I(X; Y; Z) > 0$, would indicate \emph{redundancy}, meaning $Z$ is either a mediator ($X \rightarrow Z \rightarrow Y$ or $X \leftarrow Z \leftarrow Y$) or a common cause ($X \leftarrow Z \to Y$). 
Because these latter patterns are Markov equivalent, their edge orientation cannot be established via the interaction information signatures, nonetheless, following the local orientation of already identified v-structures, Meek's rules \cite{meek1995causal} are applied to propagate the remaining undirected edges. These rules essentially prevent propagating `illegal' v-structures and enforce the acyclicity condition. For example, if the mid-node in the partially directed triple $X \rightarrow Z - Y$ was already identified not to be a collider, then $Z - Y$ must be oriented as $Z \rightarrow Y$ in order to prevent the creation of an illegal v-structure. Similarly, if orienting an edge in one direction would create a closed loop, that orientation is rejected, and the edge is oriented in the opposite direction to maintain acyclic causal flow.

Crucially, some of the edges may remain undirected even after these propagation rules are applied, representing a fundamental limit of discovering causal structure purely from observational data. As such, the algorithm preserves the undirected edges in returning the final output, a Maximal Ancestral Graph (MAG), along with directed edges ($\rightarrow$) and bi-directed edges ($\leftrightarrow $). These latter are introduced when a dependency $X-Y$ cannot be explained by any observed variable, yet the orientation signatures for the pair are contradictory. For instance if discovered v-structures and subsequent propagation rules suggest a arrowhead pointing away from both $X$ and from $Y$ in the following pattern: $\rightarrow X - Y \leftarrow$, rather than forcing a directed relationship which would either introduce illegal v-structures or create cycles, the algorithm assigns a bi-directed edge $\leftrightarrow$ to indicate that $X$ and $Y$ share a hidden (latent) parent $U$ ($\rightarrow X \leftarrow U \rightarrow Y \leftarrow$).

\section{}
\label{sec_appendix_C}

\subsection{Back-door paths in $\mathcal{G}_{\underline{V_7}}$}
\label{sec:backdoorPaths_in_V7}
The following 24 simple back-door paths are identified in the mutilated graph $\mathcal{G}_{\underline{V_7}}$. They are categorised into two groups based on their initial topological segments. Group 1: paths starting with chain segments ($V_7 \leftarrow V_8 \leftarrow \dots Y$):

\begin{enumerate}[{\arabic*:}]
  \item $(V_7, V_8, V_{13}, V_1, Y)$
  \item $(V_7, V_8, V_{13}, V_5, V_1, Y)$
  \item $(V_7, V_8, V_{13}, V_5, V_{12}, V_1, Y)$
  \item $(V_7, V_8, V_{13}, V_5, V_2, V_1, Y)$
  \item $(V_7, V_8, V_{13}, V_5, V_2, U_1, V_1, Y)$
  \item $(V_7, V_8, V_1, Y)$
  \item $(V_7, V_8, V_5, V_1, Y)$
  \item $(V_7, V_8, V_5, V_{13}, V_1, Y)$
  \item $(V_7, V_8, V_5, V_{12}, V_1, Y)$
  \item $(V_7, V_8, V_5, V_2, V_1, Y)$
  \item $(V_7, V_8, V_5, V_2, U_1, V_1, Y)$
  \item $(V_7, V_8, U_2, Y)$
\end{enumerate}
and Group 2: paths involving a collider segment ($V_7 \leftarrow V_9 \to V_8 \leftarrow \dots Y$) with $V_8$ as the mid-node:

\begin{enumerate}[{\arabic*:}]
\setcounter{enumi}{12}
  \item $(V_7, V_9, V_8, V_{13}, V_1, Y)$
  \item $(V_7, V_9, V_8, V_{13}, V_5, V_1, Y)$
  \item $(V_7, V_9, V_8, V_{13}, V_5, V_{12}, V_1, Y)$
  \item $(V_7, V_9, V_8, V_{13}, V_5, V_2, V_1, Y)$
  \item $(V_7, V_9, V_8, V_{13}, V_5, V_2, U_1, V_1, Y)$
  \item $(V_7, V_9, V_8, V_1, Y)$
  \item $(V_7, V_9, V_8, V_5, V_1, Y)$
  \item $(V_7, V_9, V_8, V_5, V_{13}, V_1, Y)$
  \item $(V_7, V_9, V_8, V_5, V_{12}, V_1, Y)$
  \item $(V_7, V_9, V_8, V_5, V_2, V_1, Y)$
  \item $(V_7, V_9, V_8, V_5, V_2, U_1, V_1, Y)$
  \item $(V_7, V_9, V_8, U_2, Y)$
\end{enumerate}

\subsection{Back-door paths in $\mathcal{G}_{\underline{V_1}}$}
\label{sec:backdoorPaths_in_V1}
The following 27 simple back-door paths are identified in the mutilated graph $\mathcal{G}_{\underline{V_1}}$. They are categorised into three groups based on their initial topological segments. Group 1: paths starting with chain segments ($V_1\leftarrow V_{12} \leftarrow V_5 \rightarrow \dots Y $):

\begin{enumerate}[{\arabic*:}]
  \item $(V_1, V_{12}, V_5, V_8, V_7, Y)$
  \item $(V_1, V_{12}, V_5, V_8, V_9, V_7, Y)$
  \item $(V_1, V_{12}, V_5, V_8, U_2, Y)$
  \item $(V_1, V_{12}, V_5, V_{13}, V_8, V_7, Y)$
  \item $(V_1, V_{12}, V_5, V_{13}, V_8, V_9, V_7, Y)$
  \item $(V_1, V_{12}, V_5, V_{13}, V_8, U_2, Y)$
  \item $(V_1, V_{12}, V_5, V_2, V_7, V_8, U_2, Y)$
  \item $(V_1, V_{12}, V_5, V_2, V_7, Y)$
  \item $(V_1, V_{12}, V_5, V_2, V_7, V_9, V_8, U_2, Y)$
\end{enumerate}
Group 2: paths involving a fork segment ($V_1 \leftarrow V_5 \rightarrow  \dots Y$) with $V_5$ as the mid-node:
\begin{enumerate}[{\arabic*:}]
\setcounter{enumi}{9}
  \item $(V_1, V_5, V_8, V_7, Y)$
  \item $(V_1, V_5, V_8, V_9, V_7, Y)$
  \item $(V_1, V_5, V_8, U_2, Y)$
  \item $(V_1, V_5, V_{13}, V_8, V_7, Y)$
  \item $(V_1, V_5, V_{13}, V_8, V_9, V_7, Y)$
  \item $(V_1, V_5, V_{13}, V_8, U_2, Y)$
  \item $(V_1, V_5, V_2, V_7, V_8, U_2, Y)$
  \item $(V_1, V_5, V_2, V_7, Y)$
  \item $(V_1, V_5, V_2, V_7, V_9, V_8, U_2, Y)$
\end{enumerate}
and Group 3: paths involving a collider segment ($V_1 \leftarrow U_1 \to V_2 \leftarrow \dots Y$) with $V_2$ as the mid-node:
\begin{enumerate}[{\arabic*:}]
\setcounter{enumi}{18}
  \item $(V_1, U_1, V_2, V_5, V_8, V_7, Y)$
  \item $(V_1, U_1, V_2, V_5, V_8, V_9, V_7, Y)$
  \item $(V_1, U_1, V_2, V_5, V_8, U_2, Y)$
  \item $(V_1, U_1, V_2, V_5, V_{13}, V_8, V_7, Y)$
  \item $(V_1, U_1, V_2, V_5, V_{13}, V_8, V_9, V_7, Y)$
  \item $(V_1, U_1, V_2, V_5, V_{13}, V_8, U_2, Y)$
  \item $(V_1, U_1, V_2, V_7, V_8, U_2, Y)$
  \item $(V_1, U_1, V_2, V_7, Y)$
  \item $(V_1, U_1, V_2, V_7, V_9, V_8, U_2, Y)$
\end{enumerate}
\end{document}